\pdfoutput=1
\documentclass[11pt]{article}

\usepackage[dvipsnames]{xcolor}

\usepackage{acl} %[review]
\usepackage{times}
\usepackage{latexsym}
\usepackage{amsthm,amsmath,amssymb}
\usepackage{mathrsfs}
\usepackage{makecell}
\usepackage[utf8]{inputenc}
\usepackage{graphicx}
\usepackage{multirow}
\usepackage{authblk}
\usepackage{bbding}
\usepackage{booktabs}
\usepackage{tabularx}
\usepackage{enumitem}
\usepackage{float}
\usepackage{subfigure}
\usepackage{tcolorbox}

\usepackage[T1]{fontenc}
\usepackage[utf8]{inputenc}

\usepackage{microtype}

\usepackage{inconsolata}

\usepackage{CJKutf8}
\title{Analyzing Temporal Complex Events with Large Language Models? \\ A Benchmark towards Temporal, Long Context Understanding}

\author[1]{\textbf{Zhihan Zhang}}
\author[2]{\textbf{Yixin Cao}\thanks{Corresponding author}}
\author[3]{\textbf{Chenchen Ye}\thanks{Work done during her work experience in National University of Singapore.}}
\author[4]{\textbf{Yunshan Ma}}
\author[5]{\textbf{Lizi Liao}}
\author[6]{\textbf{Tat-Seng Chua}}

\makeatletter
\renewcommand\AB@affilsepx{, \protect\Affilfont}
\makeatother
\affil[1,2]{School of Computer Science, Fudan University}

\makeatletter
\renewcommand\AB@affilsepx{\\ \protect\Affilfont}
\makeatother
\affil[3]{University of California, Los Angeles}

\makeatletter
\renewcommand\AB@affilsepx{, \protect\Affilfont}
\makeatother
\affil[4,6]{National University of Singapore}

\makeatletter
\renewcommand\AB@affilsepx{, \protect\Affilfont}
\makeatother
\affil[5]{Singapore Management University \authorcr \texttt{zhangzhihan22@m.fudan.edu.cn}}

\begin{document}
\maketitle
\begin{abstract}

The digital landscape is rapidly evolving with an ever-increasing volume of online news, emphasizing the need for swift and precise analysis of complex events. We refer to the complex events composed of many news articles over an extended period as Temporal Complex Event (TCE). This paper proposes a novel approach using Large Language Models (LLMs) to systematically extract and analyze the event chain within TCE, characterized by their key points and timestamps. We establish a benchmark, named TCELongBench, to evaluate the proficiency of LLMs in handling temporal dynamics and understanding extensive text. This benchmark encompasses three distinct tasks - reading comprehension, temporal sequencing, and future event forecasting. In the experiment, we leverage retrieval-augmented generation (RAG) method and LLMs with long context window to deal with lengthy news articles of TCE. Our findings indicate that models with suitable retrievers exhibit comparable performance with those utilizing long context window.

\end{abstract}

\section{Introduction}
\begin{figure}[t]
    \centering
    \includegraphics[scale=0.56]{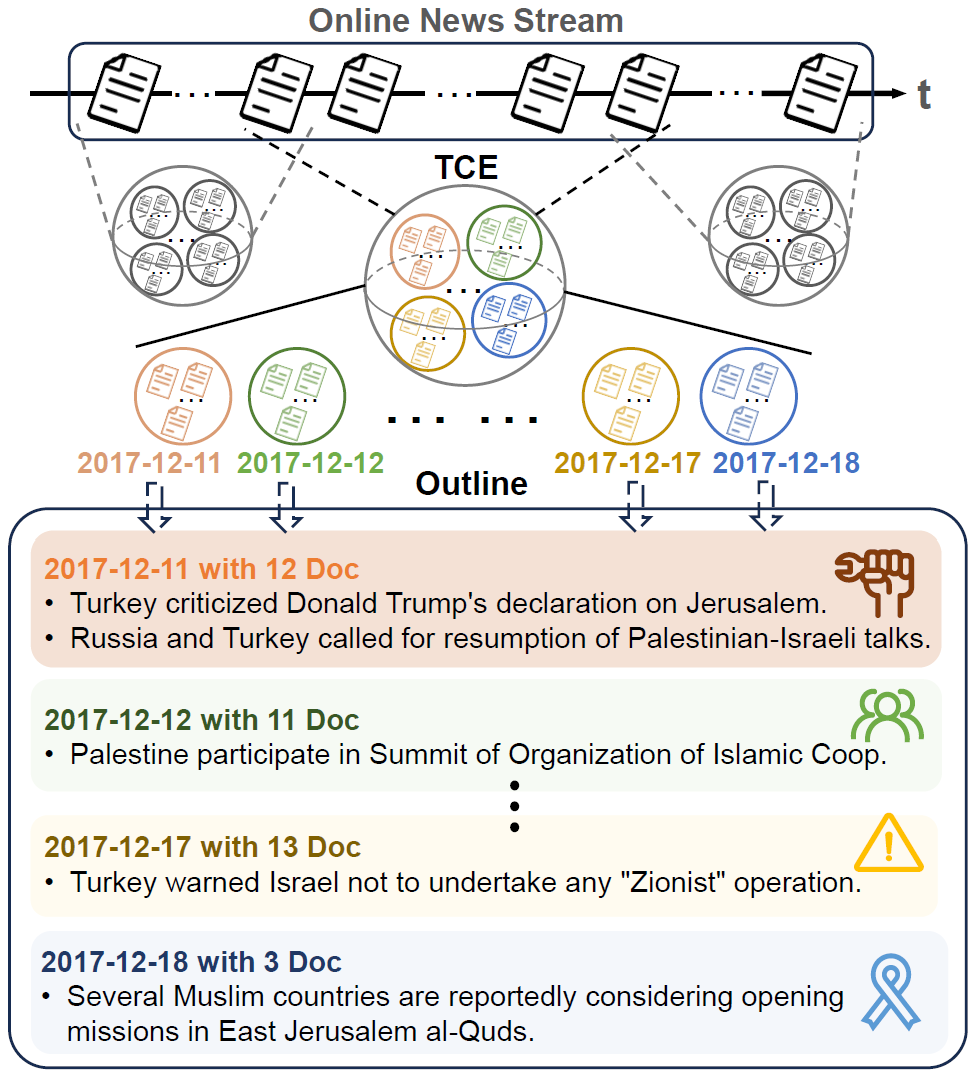}
    \caption{An example of temporal complex event (TCE) around Israeli-Palestinian conflict during December 2017. A TCE consists of many news articles with multiple timestamps.
    Our work extracts the outline of TCE. } 
    \vspace{-0.3cm}
    \label{fig:example_intro}
\end{figure} 

In today's digital age, the flood of online news highlights the urgent need for quick and precise event analysis. Prior work in topic detection has mainly clustered news articles by representation similarity to identify stories from news streams \cite{saravanakumar-etal-2021-event, 10.1145/3543507.3583507}. Extending this approach, our focus shifts to the temporal dynamics of these stories, which we term Temporal Complex Events (TCE) \cite{ma2023structured}. TCEs consist of semantically related articles that together narrate the development of various entities over time (refer to Figure \ref{fig:example_intro}). Understanding the genesis and evolution of TCE, as well as predicting future developments, holds considerable significance for meeting the practical needs of decision-makers, stakeholders, and even the general public interest.

Existing research in complex event analysis has made significant strides but is constrained by inadequate natural language processing (NLP) techniques. Some works \cite{gholipour-ghalandari-etal-2020-large, 10.1145/3543507.3583295} aims at provide concise insights into real-word events, utilizing data mining method or human-curated datasets. Another line of works \cite{li-etal-2021-future, zhu-etal-2023-diffusion} further tracks the temporal progression of complex events by converting news articles into structured data, such as temporal knowledge graphs (TKGs). The information extraction (IE) methods involved, however, tend to be costly and error-prone. Interestingly, how can modern powerful NLP models be applied to complex event analysis, and the extent to which they are aware of its temporal dynamics, remain challenging to determine.

In this paper, inspired by the extensive success of LLMs across various NLP challenges, we delve into their suitability for analyzing TCEs and assess their prowess in understanding temporal and long contexts. First, LLMs typically have a limitation in input length, e.g. 4,096 tokens, while a TCE may span tens of news articles and then tens of thousands of tokens (i.e., an average of 29 articles and 18,589 tokens in our experimental datasets). Even if longer context window enables LLMs to take in all articles, existing works~\cite{bai2023longbench, xu2024retrieval} have demonstrated their inferior performance with lengthy context. 
Second, LLMs, pre-trained for next token prediction, sometimes fall short in temporal reasoning tasks \cite{tan2023benchmarking}.
For TCE analysis, this limitation becomes apparent as it necessitates precise event-timestamp correlation and a deep understanding of chronological and causal connections. Furthermore, building on top of lengthy past events and their temporal relations, their potential for predicting future events is still under-explored.

To this end, we propose a LLM-based pipeline for TCE outline extraction, and build a large-scale benchmark \textbf{T}CE\textbf{L}ong\textbf{B}ench (TLB) for comprehensive investigation. Inspired by \cite{reddy2023smartbook}, we aim at providing a coherent and chronological representation of TCE, i.e. outline with a timeline. We apply a hierarchical summarization framework and then leverage LLM's in-context learning (ICL) ability \cite{NEURIPS2020_1457c0d6} to extract key points on each day, in the form of sentences. After de-duplication, key points across all timestamps constitute the outline of TCE. 

Based on these, we build TCELongBench for temporal, long context evaluation. It contains 88,821 question answering (QA) pairs from 2,289 TCEs, tailored to three distinct tasks: \textit{TLB-detail} QA, which tests LLMs' ability to find evidence across numerous articles; \textit{TLB-order} QA, focusing on understanding temporal sequences; and \textit{TLB-forecast} QA, challenging LLMs to predict future events based on past information. To ensure dataset integrity, we employed a \textit{generate-then-verify} paradigm, leading to a dataset with an 88\% quality rating across human evaluation metrics.

In our analysis, we employed both retrieval-augmented generation (RAG) methods and LLMs optimized for long contexts to navigate the extensive narratives typical of TCEs. Our findings reveal that (1) while retrievers are crucial for RAG methods, their effectiveness is variable; (2) long-context models excel in managing long temporal sequences but may lead to inferior performance; and (3) models equipped with apt retrievers can match the performance of those designed for long contexts. To sum up, our contributions are threefold:

\begin{itemize}[leftmargin=*, itemsep=1pt, topsep=1pt, parsep=1pt]
    \item We leverage LLMs to extract the outlines and form event chains of TCEs.
    \item We build TCELongBench that consists of three tasks aiming at testing the model's capability of temporal, long text understanding.
    \item We conduct extensive experiments of LLMs leveraging RAG method and LLMs with long context window.
\end{itemize}

\section{Related Work}

\noindent{\textbf{Complex Event Analysis.}}
Some works around complex event analysis rely on schema to extract temporal knowledge graphs from narratives, such as IED \cite{li-etal-2021-future} and RESIN-11 \cite{du-etal-2022-resin}. To further capture the temporal characteristics of complex events, \citet{ma2023structured} contribute MidEast-TE that associates each event with a timestamp. However, their intricate information extraction pipelines are time-consuming and may lead to unexpected errors for event analysis. Several studies also explore the unstructured storyline of complex events from multiple documents, in the form of summaries \cite{gholipour-ghalandari-etal-2020-large}, timeline \cite{steen-markert-2019-abstractive, gholipour-ghalandari-ifrim-2020-examining} and event mentions \cite{10.1145/3543507.3583295}. In this paper, we extract outlines from TCEs, consisting of key points (sentences) that record the detailed actions of entities with suitable granularity and unfold the whole story within the TCE over time.

A more recent work \cite{reddy2023smartbook} formulates a report generation task around complex events using LLMs, but falls short in large-scale datasets and quantitative analysis on the report quality. However, before delving into long text generation, we aim at evaluating the LLM's capability of understanding temporal, long text in TCE, and contribute a QA dataset for quantitative comparisons of various baselines.

\noindent{\textbf{Related Benchmarks.}} There are two strands of benchmarks related to TCELongBench. First, temporal reasoning benchmarks \cite{zhang-choi-2021-situatedqa, dhingra-etal-2022-time, tan2023benchmarking} mostly focus on Event-Time, Event-Event and/or Time-Time relations of chronicles in Wikipedia. For example, TRAM \cite{wang2023tram} encompasses ten temporal reasoning tasks, including temporal ordering without any context. ForecastQA \cite{jin-etal-2021-forecastqa} are proposed to develop methods for event forecasting with large volumes of unstructured text data. Second, long text understanding benchmarks \cite{bai2023longbench, dong2023bamboo, an2023leval, shaham2023zeroscrolls} aim at evaluating long text modeling with multiple tasks, such as summarization, question answering, code completion, etc. In contrast, TCELongBench evaluates the model's understanding of TCEs from three tasks, requiring temporal reasoning, long text understanding as well as forecasting abilities.

\section{Task Definition}

\begin{figure*}[t]
    \centering
    \includegraphics[scale=0.75]{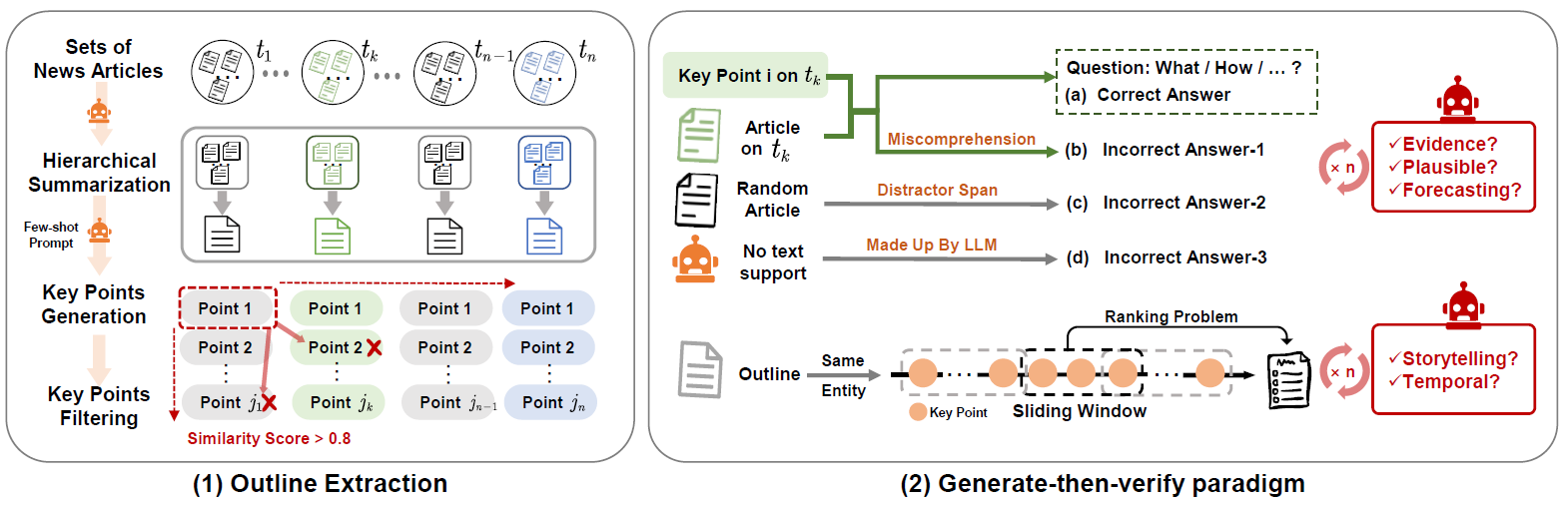}
    \vspace{-0.2cm}
    \caption{Pipeline of outline extraction and generate-then-verify paradigm.}
    \vspace{-0.4cm}
    \label{fig:pipeline_dataset}
\end{figure*}

Existing work has identified TCEs from news articles by clustering their semantic embeddings concatednated with temporal indexes \cite{ma2023structured}.
Each TCE has $n$ timestamps, i.e. a timeline $\mathcal{T} = \{t_k: k \in [1,n] \}$, and news articles $\mathcal{A}_{n} = \{\mathrm{A_k}: k \in [1,n] \}$, where $\mathrm{A_k}$ is the set of news articles on $t_k$.
On each timestamp $t_k$, we extract $j_k$ number of key points from $\mathrm{A_k}$, expressed as $\mathrm{P_k}=\{P_{1,k},\dots,P_{j_k,k}\}$. Each key point is a concise and informative sentence. The collection of key points across all timestamps forms the TCE's outline $\mathcal{P} = \{\mathrm{P_k}: k \in [1,n] \}$. Note that news articles accessible to models are $\mathcal{A}_{n-1} =\{\mathrm{A_k}: k \in [1,n-1] \}$ in our experiment as $\mathrm{A_n}$ is used for generating forecasting questions.

\noindent{\textbf{TLB-detail.}} This is a reading comprehension task aiming at testing the model's \textit{ability to locate and understand detailed information across numerous articles.} The input is a question $Q$, a set of shuffled choices $\mathrm{C} = \{C_r:r\in[1,4]\}$, and $\mathcal{A}_{n-1}$, while the output is a choice $C_l \in \mathrm{C}$.

\noindent{\textbf{TLB-order.}} This is an ordering task aiming at testing a model's \textit{ability to capture the event-event relations across timestamps}. The input is a set of shuffled choices $\mathrm{C} = \{C_r:r\in[1,R]\}$ and $\mathcal{A}_{n-1}$, while the output is the chronological order of the choices $\{C_{O_1},\dots,C_{O_R}\}$.

\noindent{\textbf{TLB-forecast.}} This is a forecasting task aiming at testing a model's ability to \textit{predict future event given historical data}. We have two settings of answering forecasting questions, multi-choice and open-domain.
In multi-choice setting, the input is a question $Q$, a set of shuffled choices $\mathrm{C} = \{C_r:r\in[1,4]\}$ and $\mathcal{A}_{n-1}$; the output is a choice $C_l \in \mathrm{C}$. In open-domain setting, we only have question $Q$ and $\mathcal{A}_{n-1}$ as the input, while the output is open for LLMs.

For each question in TLB-detail and TLB-forecast, the text span that supports its correct answer lies in the gold article $A_{gold}$ on $t_{gold}$. While $A_{gold}$ in TLB-detail follows $A_{gold} \in \mathcal{A}_{n-1}$, the $A_{gold}$ in TLB-forecast is within $\mathrm{A_n}$, not accessible during evaluation.
Moreover, articles on $t_{gold}$ except $A_{gold}$ may offer supporting evidence to the correct answer, suggesting that identifying $t_{gold}$, rather than precisely matching $A_{gold}$, is also pivotal in determining the correct answer.

\section{Outline Extraction}

Inspired by \citet{10.1145/3543507.3583295} and \citet{rashkin-etal-2020-plotmachines}, we propose a LLM-based outline extraction pipeline, which tersely organizes the primary content of TCEs along with a clear timeline. Outline in our work consists of key points from all timestamps, each of which is a concise and informative sentence. These key points represent TCEs with suitable granularity, recording the detailed actions of entities and unfolding the whole story over the timelines. Neither the fine-grained TKG nor event mention (phrase) could capture the intricate relations of multiple entities within TCEs. 

Our LLM-based outline extraction pipeline consists of three parts, summarization, key point generation and key point filtering (Figure \ref{fig:pipeline_dataset} (1)). 
Initially, we implement a hierarchical summarization framework to filter out extraneous peripheral events, using \texttt{xgen-7b-8k-inst} \cite{nijkamp2023xgen7b}. This framework operates as follows: on each timestamp $t_k$, we summarize each news article within $\mathrm{A_k}$ to distill their essential contents, and then summarize these articles' summaries to obtain the central event on $t_k$. Consequently, we compile the daily summaries across all timestamps as $\mathrm{S} = \{S_k: k \in [1,n] \}$.

We then leverage LLM's ICL ability to  partition daily summaries into key points. We design a few-shot prompt (Table \ref{table:prompt_point_extr}), and ask \texttt{gpt-3.5-turbo-instruct} to generate key points $\mathrm{\hat{P}_k}=\{\hat{P}_{1,k},\dots,\hat{P}_{\hat{j}_k,k}\}$ given a daily summary $S_k$. Instructions in the prompt specify that key points should be independent, concise, and comprehensive, avoiding any pronoun. Moreover, the prompt incorporates three human-curated examples to steer the model to better performance.

Finally, we implement a filtering mechanism to enhance the quality of timeline. We eliminate redundant key points that duplicate previously conveyed information, by calculating two similarity scores using \texttt{sup-simcse-bert} \cite{gao2021simcse} and \texttt{quora-distilroberta} \cite{reimers-2020-multilingual-sentence-bert}.
If any of the similarity scores between $P_{i,m}$ on $t_m$ and $P_{j,k}$ on $t_k$ exceeds predefined thresholds, i.e. 0.8, we discard the key point in later position, i.e. $P_{i,m}$, since $t_m > t_k$ or $i > j$ if $t_m = t_k$. Subsequently, we obtain the TCE's outline $\mathcal{P} = \{\mathrm{P_k}: k \in [1,n] \}$.

\section{Dataset Generation and Analysis}

Based on our extracted outlines, we construct QA datasets in TCELongBench, under a \textit{generate-then-verify} paradigm. We also show the summary statistics and human evaluation results.

\subsection{Generate-then-verify Paradigm}
\label{qa_gen}

We generate questions and answers given key points and news articles, and then verify their quality from multiple aspects, including \textit{Evidence}, \textit{Plausible}, \textit{Forecasting}, \textit{Storytelling} and \textit{Temporal}. 

\subsubsection{TCE QA Generation}

\textbf{TLB-detail} and \textbf{TLB-forcast} are in the form of multi-choice question answering (MCQ). We leverage LLM and follow the STARC annotation framework \cite{berzak-etal-2020-starc} to generate question and misleading choices. In specific, for question generation, we ask \texttt{gpt-3.5-turbo-instruct} to propose a question along with its correct answer for each key point in $\mathcal{P}$ . Here we adopt a few-shot prompt (see Table \ref{table:prompt_detailQA_gen} and \ref{table:prompt_foreQA_gen}), where examples are from OneStopQA \cite{berzak-etal-2020-starc} and ForecastQA \cite{jin-etal-2021-forecastqa}. For misleading choices generation, we design instructions under STARC annotation framework: (1) the first choice represents a plausible misunderstanding of the article $A_{i,k}$; (2) the second one is anchored in another random article with a different timestamp $A_{\hat{i},\hat{k}}$ ($\hat{k} \neq k$), plausible to the question but incorrect; (3) the third one is made up by LLMs (see Table \ref{table:prompt_choices_gen}). Additionally, since real-world future events are not confined by candidate choices, we adopt an open-domain setting in TLB-forecast, where only questions and news articles are provided.

\textbf{TLB-order} is in the form of ranking problem. To ensure the choices to be ordered have a strong relation with each other, we formulate ranking problems by selecting the key points associated with a common entity, inspired by \citet{lin-etal-2021-conditional}. In specific, we use spaCy \cite{spacy2} to extract the entities in each key point, and then collect those sharing at least one common entity. For each common entity $e_k$ that links a branch of key points, we select every three of them with neighboring timestamps to form a ranking problem. Note that the choices in all three tasks are randomly shuffled after generation.

\begin{table*}[h!] 
\tiny
\setlength\tabcolsep{2pt}
\centering
\renewcommand{\arraystretch}{0.8}
%\scalebox{0.8}{
\resizebox{\linewidth}{!}{
\begin{tabular}{p{0.8cm}<{\centering} | p{10cm}}
 \specialrule{0.08em}{3pt}{3pt}
 \multirow{5}*{\makecell[c]{\textbf{TLB-}\\ \textbf{detail}}} & \textbf{Q:} What was Syria's response to the US's recognition of the Golan Heights as Israeli territory? \\
 & \textbf{A.} Requested UN funding to rebuild after the war. \textbf{B.} Declare military victory over ISIS in response. \\
 & \textbf{C.} Consider taking military action against Israel. \textbf{D.} \textcolor{PineGreen}{Request an urgent meeting with UN Security Council}. \\
 \specialrule{0em}{1pt}{1pt}
 & \textbf{Reasoning Path:} Syria has \textcolor{PineGreen}{asked the UN Security Council on Tuesday to hold an urgent meeting} on the US decision to recognize the Golan Heights as Israeli territory on 2019-03-38. (\textbf{Evidence of Choice D}) The correct answer is D. \\
 \specialrule{0.04em}{2pt}{2pt}
 \multirow{9}*{\makecell[c]{\textbf{TLB-}\\ \textbf{order}}} & \textbf{A.} Syria \textcolor{Orange}{requested an urgent meeting at the United Nations Security Council} to discuss US President Donald Trump's decision to recognize the Golan Heights as Israeli territory, which conflicts with UN resolutions. \\
 & \textbf{B.} Lebanese government states that \textcolor{Purple}{Shebaa Farms were not part of Golan Heights} as Israel did not annex their territory. \\
 & \textbf{C.} \textcolor{Maroon}{The US maps} will be redrawn to include the Golan Heights as a part of Israel. \\
  \specialrule{0em}{1pt}{1pt}
 & \textbf{Reasoning Path:} Syria has \textcolor{Orange}{asked the UN Security Council to hold an urgent meeting} on 2019-03-28. (\textbf{Evidence of Choice A}) A Lebanese official claims that \textcolor{Purple}{Shebaa Farms were not part of the Golan Heights} because “no one mentioned our land to declare its annexation to Israel” on 2019-03-31. (\textbf{Evidence of Choice B}) \textcolor{Maroon}{The US maps} are slated to reflect Donald Trump’s recognition of Israeli sovereignty over the Golan Heights on 2019-03-29. (\textbf{Evidence of Choice C}) Following the timestamps, the correct answer is A,C,B. (\textbf{Temporal Ordering}) \\
 \specialrule{0.04em}{2pt}{2pt}
 \multirow{12}*{\makecell[c]{\textbf{TLB-}\\ \textbf{forecast}}} & \textbf{Q:} What will be the response of \textcolor{NavyBlue}{international community} to Israel's annexation of Golan Heights after 2019-04-17? \\
 & \textbf{A.} Remain silent on the issue, as they have no interest in the Middle East conflict. \\
 & \textbf{B.} Take military action against Israel, as they see their actions as a threat to global security. \\
 & \textbf{C.} Support Israel's actions and recognize their right to claim the Golan Heights as their own. \\
 & \textbf{D.} \textcolor{NavyBlue}{Condemn Israel's actions and reaffirm their stance} that the Golan Heights is not a part of Israel's sovereignty. \\
  \specialrule{0em}{1pt}{1pt}
 & \textbf{Reasoning Path:} Donald Trump’s recognition of Israeli sovereignty over the Golan Heights was \textcolor{NavyBlue}{condemned by France, Germany, UK, Russia, Syria and other countries} on 2019-03-29. \textcolor{NavyBlue}{EU also rejected} to recognize Israeli sovereignty over Syrian Golan Heights on 2019-04-16. (\textbf{Context Location}) The international community could be represented by the countries and EU mentioned in the context. (\textbf{Bridge Entity}) Given their past positions on Israel's annexation of Golan Heights, the correct answer is most likely to be D. (\textbf{Inferring based on past events}) \\ 
 \specialrule{0.08em}{2pt}{2pt}
\end{tabular}
}
\vspace{-0.3cm}
\caption{Examples of three QA tasks in TCELongBench from TCE 2762. 
}
\vspace{-0.3cm}
\label{table:analysis_QA_example}
\end{table*}

\subsubsection{TCE QA Verification}

Although powerful, LLMs may still produce illogical question or hallucination. To filter out noisy QA pairs, we perform an additional verification step as follows. For TLB-detail QA, we consider two aspects: 
\begin{itemize}[leftmargin=*, itemsep=1pt, topsep=1pt, parsep=1pt]
    \item \textit{Evidence}. Considering the quality of question and correct answer, we check if there is direct evidence in $A_{i,k}$ that supports the correct answer (see Table \ref{table:prompt_veri_evidence}).
    \item \textit{Plausible}. Considering the quality of misleading choices, we check if they are different from but sharing similar wording with the correct answer.
\end{itemize}
TLB-forecast QA further adds one aspect:
\begin{itemize}[leftmargin=*, itemsep=1pt, topsep=1pt, parsep=1pt]
    \item \textit{Forecasting} \cite{jin-etal-2021-forecastqa}. Considering the logic behind predicting future event, we check if it is true that while the question cannot be answered with certitude using historical data, it remains tractable and guessable for individuals with expertise?
\end{itemize}
For TLB-order QA, we focus on other two aspects:
\begin{itemize}[leftmargin=*, itemsep=1pt, topsep=1pt, parsep=1pt]
    \item \textit{Storytelling}. Considering the relations between choices, we check if they are connected by related entities and hopeful to form a storyline?
    \item \textit{Temporal}. Considering the time-sensitive feature of temporal ordering, we check if each choice represent an event that just happened, instead of static or past event?
\end{itemize}

Specifically, \textit{Evidence} is examined right after the question is generated, and the generation will stop if there is no supportive evidence found. For \textit{Plausible}, we keep the QA pair if its misleading choices have less-than-ten-words differences with the correct one and do not repeat it, checked by similarity scores. Moreover, we ask \texttt{gpt-3.5-turbo-instruct} to check the resting three aspects in the multi-choice QA format, \texttt{A} for passing, \texttt{B} for failing, and \texttt{C} for not knowing. Inspired by \citet{jin-etal-2021-forecastqa}, we repeat \textit{three rounds} on the same QA pair, which is qualified only when more than two rounds choose \texttt{A}.

After verification, there is a filtering procedure for dropping the repeated QA pairs. We again use the similarity and duplication scores to discard redundant questions in TLB-detail and TLB-forecast, while for TLB-order, the sets of choices that share more than one common key point will be discarded (see Appendix \ref{ap:point_dupli} for details).

\subsection{Dataset Analysis}

\noindent{\textbf{Corpus.}} We use Mideast-TE \cite{ma2023structured} corpus that has identified TCEs from GDELT.
We filter out those TCEs whose time span is too long (i.e., one month) or too short (i.e., five days). This results in 2,289 TCEs in total where average articles and days are 29.31 and 17.44 respectively.

\begin{figure}[htp]
\centering 
\vspace{-0.3cm}
\subfigure[Day Gap]{
\includegraphics[scale=0.19]{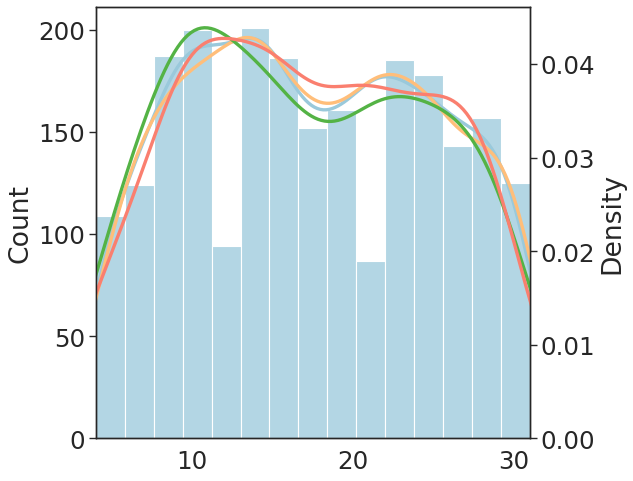}}
\subfigure[Token]{
\includegraphics[scale=0.19]{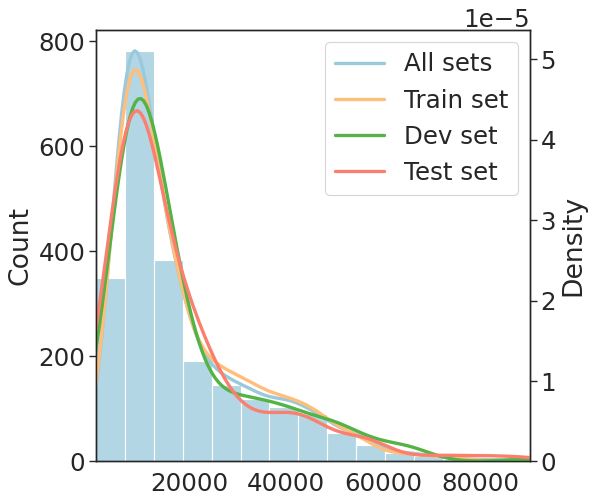}}
\vspace{-0.3cm}
\caption{Distributions of day gaps (a) and number of tokens (b). Histograms are with the left y-axis and lines of kernel density estimation are with the right y-axis.}
\vspace{-0.3cm}
\label{fig:distri}
\end{figure}

\noindent{\textbf{Statistics.}} We randomly assign TCEs into training, development and test sets following 75/15/15 proportions, shown in Table \ref{table:split_stat_1}. While the day gaps of TCE are evenly distributed within 30 days, their numbers of tokens present right-skewed distributions around  10,000 (see Figure \ref{fig:distri}).

\begin{table}[h!]
\centering
\scalebox{0.75}{
\begin{tabular}{l r c r c r c } 
\specialrule{0.08em}{3pt}{3pt}
 \multirow{2}*{\textbf{Dataset}} & \multicolumn{2}{c}{\textbf{Train}} & \multicolumn{2}{c}{\textbf{Dev}} & \multicolumn{2}{c}{\textbf{Test}} \\
  & Num. & \% & Num. & \% & Num. & \%  \\
 \hline
 \textit{Complex Event} & 1602 & 70.0  & 343 & 15.0 & 344 & 15.0  \\
 TLB-detail & 43,336 & 71.0  & 8,916 & 14.6 & 8,801 & 14.4  \\
 TLB-order & 15,149 & 71.6  & 3,048 & 14.4 & 2,967 & 14.0 \\ 
 TLB-forecast & 4,565 & 69.1  & 1,027 & 15.6 & 1,012 & 15.4 \\
\specialrule{0.08em}{3pt}{3pt}
\end{tabular}}
\caption{Numbers and proportions of TCE and QA pair in train/dev/test sets.}
\vspace{-0.3cm}
\label{table:split_stat_1}
\end{table}

\begin{figure}[htp]
    \centering
    \includegraphics[scale=0.5]{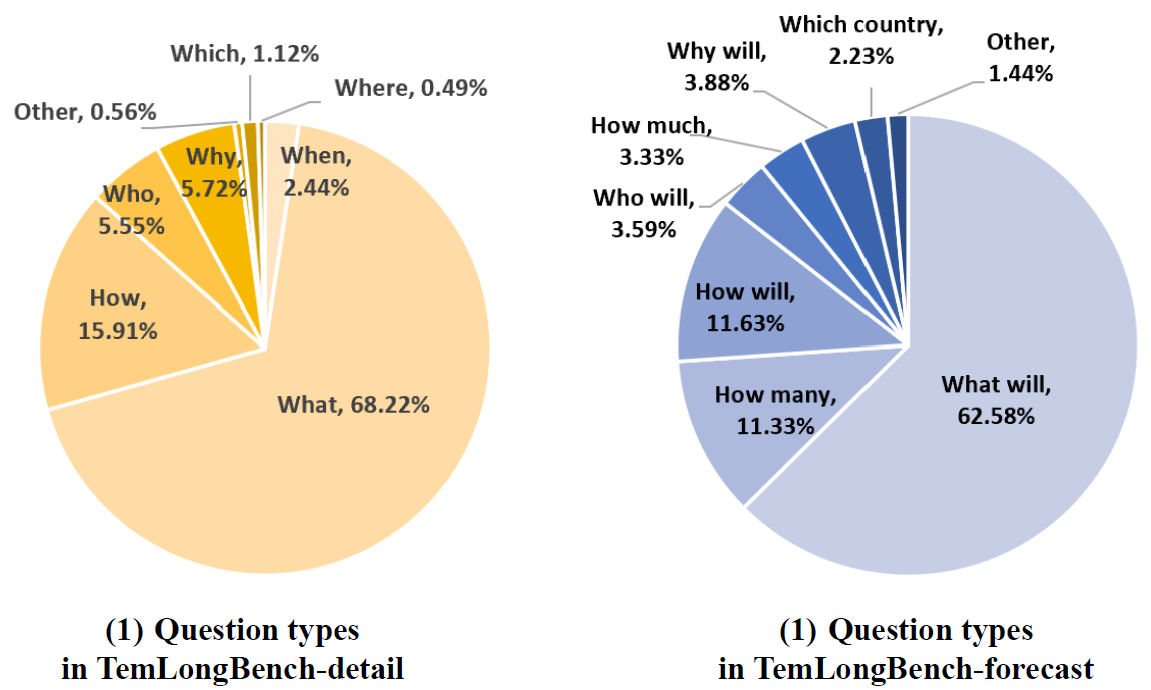}
    \caption{Question types in TLB-detail and TLB-forecast.}
    \label{fig:starts_ques}
\end{figure}

There are different question types in TLB-detail and TLB-forecast(see Figure \ref{fig:starts_ques}). MCQs in TLB-detail starts with What (68.22\%), How (15.91\%), Who (5.55\%), etc., while those in TLB-forecast starts with What will (62.58\%), How will (11.63\%), How many (11.33\%), etc. Besides, following \citet{jin-etal-2021-forecastqa}, forecasting questions end with a timestamp like "in/after/by 2019-09-18". For TLB-order QA, average day gap of choices is 5.79 days.

\noindent{\textbf{Challenges.}} As shown in Table \ref{table:analysis_QA_example}, TLB-detail requires accurately identifying relevant text spans and correlating them with candidate choices for answering reading comprehension questions. TLB-order poses a heightened challenge, involving the identification of multiple contexts with varying timestamps and linking temporal information with choices to establish their relations. TLB-forecast entails additional reasoning steps, including entity bridging and inference from historical events.

\subsection{Human Evaluation}

We ask three annotators to evaluate the quality of QA pairs in TCELongBench from multiple dimensions similar to verification step during dataset construction. The evaluation is conducted on a random sample with size 84 from ten TCEs.

Each annotator decides whether or not a QA pair satisfies one dimension by rating it with 1 or 0, 1 for meeting and 0 for failing. On average, the accuracy score of annotators over three tasks is 77.38\%, suggesting that tasks in our TCELongBench are quite challenging for humans. Moreover, the evaluation results are 97.61\% for \textit{Context}, 86.90 \% for \textit{Evidence}, 95.67\% for \textit{Reasonable}, 90.12\% for \textit{Plausible}, 77.78\% \textit{Temporal} and 95.56\% for \textit{Storytelling} (see Appendix \ref{ap:hu_eval} for definitions of each dimension). This result proves the high-quality of TCELongBench, which are mainly attributed to two elaborate procedures during dataset construction: (1) few-shot prompts with detailed instructions and human-curated examples from existing datasets \cite{berzak-etal-2020-starc, jin-etal-2021-forecastqa}; (2) multi-turn verification by LLMs.

\section{Benchmarking Experiments}

\begin{table*}[h!] \small
\setlength\tabcolsep{3pt}
\centering
\renewcommand{\arraystretch}{0.95}
\scalebox{1}{
\begin{tabular}{l | c | c c c c c c c c c c}
 \toprule[0.8pt]
 \multirow{3}*{\textbf{Model}} & \multirow{3}*{\makecell[c]{\textbf{Retriever} \\ /\textbf{Length}}} & & \textbf{TLB-detail}  & & \multicolumn{3}{c}{\textbf{TLB-order}} & & \multicolumn{3}{c}{\textbf{TLB-forecast}} \\
 & & & \multirow{2}*{Acc.} & & \multirow{2}*{Acc. $\Uparrow$} & \multirow{2}*{F1 $\Uparrow$} & \multirow{2}*{Dist $\Downarrow$} & & MCQ & \multicolumn{2}{c}{Open-domain} \\
 & & & & &  &  &  & & Acc. & BLEU & METEOR \\ % \hline
 \specialrule{0.08em}{1.5pt}{1.5pt}
 \multirow{4}*{\texttt{vicuna-7b-4k}} & w/o context & & 26.3 & & 12.2 & 24.0 & 2.07 &  & 26.8 & 0.89 & 19.3   \\
 & BM25 & & 68.3 & & 12.9 / 13.2 & 25.4 / 25.3 & 2.02 / 2.02 &   & 46.6 & 1.20 & 22.2 \\
 & Openai & & 68.5 & & 12.3 / 13.0 & 24.2 / 25.6 & 2.06 / 2.00 &  & 48.2 & 1.13 & 22.5 \\
 & Hybrid & & 68.6 & & 13.2 / 14.1 & 26.1 / 27.0 & 1.99 / 1.96 &  & 48.3 & 1.36 & \underline{22.8}  \\
 \specialrule{0.04em}{1pt}{1pt}
 \multirow{4}*{\texttt{Llama-2-7b-4k}} & w/o context & & 25.3 & & 9.3 & 18.2 & 2.29 &  & 15.6 & 0.65 & 18.8  \\
 & BM25 & & 70.6 & & 11.1 / 12.9 & 22.5 / 24.2 & 2.13 / 2.09 &  & 48.6 & 1.10 & 21.5 \\
 & Openai & & 68.2 & & 10.8 / 12.3 & 22.1 / 23.4 & 2.14 / 2.11 &  & 49.6 & 0.93 & 21.6 \\
 & Hybrid & & 69.2 & & 11.4 / 14.5 & 22.5 / 26.4 & 2.13 / 2.00 &  & 49.1 & 0.99 & 21.9  \\
 \specialrule{0.04em}{1pt}{1pt}
 \multirow{4}*{\texttt{vicuna-13b-4k}} & w/o context & & 34.7 & & 17.8 & 34.7 & 1.66 & & 30.9 & 0.82 & 18.6  \\
 & BM25 & & 72.4 & & 15.7 / 18.6 & 30.8 / 33.9 & 1.80 / 1.72 & & 43.4 & 1.28 & 22.4 \\
 & Openai & & 71.5 & & 16.4 / 18.8 & 31.0 / 33.7 & 1.80 / 1.72 &  & 42.2 & 1.23 & 22.5 \\
 & Hybrid & & 75.3 & & 14.7 / 19.0 & 28.3 / 34.5 & 1.90 / 1.69 &  & 40.7 & 1.20 & 22.5 \\
 \specialrule{0.04em}{1pt}{1pt}
 \multirow{4}*{\texttt{Llama-2-13b-4k}} & w/o context & & 35.2 & & 18.3 & 33.8 & 1.67 & & 29.2 & 0.42 & 16.6  \\
 & BM25 & & 78.2 & & 10.5 / 15.4 & 20.4 / 25.6 & 2.21 / 2.05 & & 58.4 & 1.01 & \underline{22.8} \\
 & Openai & & 76.5 & & 9.0 / 16.7 & 16.9 / 27.4 & 2.33 / 2.00 &  & 59.2 & 0.97 & 22.6 \\
 & Hybrid & & 79.8 & & 10.1 / 14.8 & 20.0 / 25.4 & 2.22 / 2.06 & & 57.2 & 0.90 & 22.6  \\
 \specialrule{0.04em}{1pt}{1pt}
 \multirow{4}*{\texttt{gpt-3.5-4k}} & w/o context & & 56.5 & & 16.8 & 33.2 & 1.67 & & 54.2 & 1.25 & 17.7   \\
 & BM25 & & 81.8 & & 15.4 / 18.1 & 29.1 / 32.2 & 1.87 / 1.81 &  & 57.7 & 1.71 & 21.0 \\
 & Openai & & 81.9 & & 14.8 / 18.3 & 27.7 / 32.2 & 1.93 / 1.80 & & 58.0 & 1.64 & 21.4 \\
 & Hybrid & & \underline{84.0} & & 15.3 / 18.8 & 28.1 / 32.4 & 1.91 / 1.80 &  & \underline{61.7} & \textbf{2.89} & 21.5  \\
 \specialrule{0.04em}{2pt}{2pt}
 \texttt{vicuna-7b-16k} & \multirow{3}*{16k} & & 37.3 & & 15.3 & 30.8 & 1.80 & & 37.9 & 1.55 & \textbf{23.4} \\
 \texttt{longchat-7b-16k} & & & 34.4 & & 9.7 & 18.5 & 2.27 & & 30.0 & 1.05 & 19.8 \\
 \texttt{gpt-3.5-16k} & & & 82.4 & & 19.5 & 33.9 & 1.75 & & 61.4 & \underline{1.79} & 21.9 \\
 \specialrule{0.04em}{2pt}{2pt}
\texttt{longchat-7b-32k} & \multirow{2}*{32k} & & 26.5 & & 8.5 & 17.1 & 2.33 & & 22.2 & 1.33 & 22.5 \\
 \texttt{chatglm3-6b-32k} & & & 79.4 & & \underline{19.8} & \underline{35.4} & \underline{1.64} & & 60.3 & 1.11 & 14.6 \\
 \specialrule{0.04em}{2pt}{2pt}
 \texttt{gpt-4-128k} & 128k & &  \textbf{91.9}$^\ast$ & & \textbf{29.6} & \textbf{45.0} & \textbf{1.42} & & \textbf{72.0} & 1.06 & \textbf{23.4} \\
 \specialrule{0.08em}{1.5pt}{1.5pt}
 % \bottomrule[0.8pt]
\end{tabular}
}
\vspace{-0.2cm}
\caption{Results of TCELongBench. For retrievers, w/o context means answering without any retrieved context; BM25, Openai and Hybrid represent sparse, dense and hybrid retrievers respectively. For TLB-order, “number1/number2” is the result of \textit{Retrieve Once} strategy and \textit{Retrieve One by One} strategy respectively. $^\ast$ means experimenting on a random sub-sample with size 1,000, due to cost limitation.}
\vspace{-0.3cm}
\label{table:all_exp}
\end{table*}

\subsection{Comparing Models}
\begin{figure}[htp]
    \centering
    \includegraphics[scale=0.65]{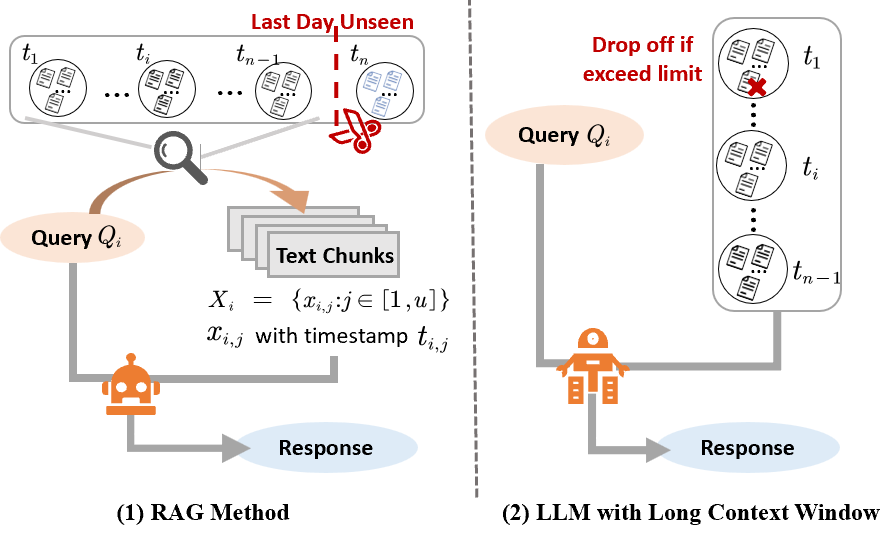}
    \caption{Evaluation pipeline of models using RAG method and LLM with Long Context Window.}
    \label{fig:pipeline_models}
\end{figure}

We apply RAG method and LLMs with long context window to our experiments (see Figure \ref{fig:pipeline_models}). Moreover, we conduct evaluation on both LLMs and retrievers.

\noindent{\textbf{RAG Method.}} LLMs with short context window (4,096 tokens) are able to read long text with the help of retrievers. We use four open-source chat models with two sizes (\texttt{vicuna-7b-4k}, \texttt{vicuna-13b-4k}, \texttt{Llama-2-7b-4k} and \texttt{Llama-2-13b-4k}) and one close-source model (\texttt{gpt-3.5-4k}).
As for retrievers, we experiment with a sparse retriever BM25, a dense retriever based on \texttt{text-embedding-ada-002} and a hybrid retriever combining the former two retrievers with a re-ranker. We set the number of retrieved text chunks $u$ and its size $l$ to be 3 and 512 respectively, considering the content window limit.

\noindent{\textbf{LLM with Long Context Window.}} Recent studies have committed to enhancing the long text modeling techniques of LLMs, extending the context length to 16k, 32k and even 128k. In our experiments, we use three models with 16k context length (\texttt{vicuna-7b-16k}, \texttt{longchat-7b-16k} and \texttt{gpt-3.5-16k}), two models with 32k (\texttt{longchat-7b-32k} and \texttt{chatglm3-6b-32k}), and one model with 128k ( \texttt{gpt-4-128k}). All accessible news articles within TCE along with their timestamps and the QA pair are fed into their context window. However, if the number of tokens exceeds the input limit, we discard the articles from $t_{n-1}$ in TLB-detail and TLB-order, and from $t_1$ in TLB-forecast, except those on the gold timestamp. Please see Appendix \ref{ap:inpt_limit} for more details.

\subsection{Evaluation Metrics} 
\noindent{\textbf{Task Evaluation.}} For MCQ in TLB-detail and TLB-forecast, we evaluate using Accuracy. In TLB-order, it is evaluated by Accuracy, weighted F1 score, and Levenshtein distance \cite{lev_dist}. For the open-domain setting in TLB-forecast, we evaluate using BLEU \cite{10.3115/1073083.1073135} and METEOR \cite{banerjee-lavie-2005-meteor}.

\noindent{\textbf{Retriever Evaluation.}} We evaluate the retriever's ability to locate the gold articles and timestamps. In TLB-detail, we use two metrics: (1) \textit{Acc\_Doc} measures the ratio of questions in which the retriever finds the gold articles; and (2) \textit{Acc\_Date} measures the ratio of questions in which the retriever finds the gold timestamps. 
In TLB-order, a ranking problem consist of three shuffled key points as choices, each having a timestamp. So its evaluation metric \textit{Acc\_Dates} measures the ratio of ranking problems in which the retriever locates all three timestamps of choices. Please see Appendix \ref{ap:retri_eval} for more details and math formulas.

Prompts templates for evaluation are in Appendix \ref{ap:prompt_eval}, following "[System Message] [Context] Given above articles, please answer the question. [Question] [Candidate Choices]" pattern.

\begin{figure*}[t]
\centering
\vspace{-0.35cm}
\subfigtopskip=2pt
\subfigbottomskip=2pt
\subfigcapskip=-5pt
\subfigure[Input Length]{
    \label{fig:distr_long_qa_acc}
    \includegraphics[scale=0.38]{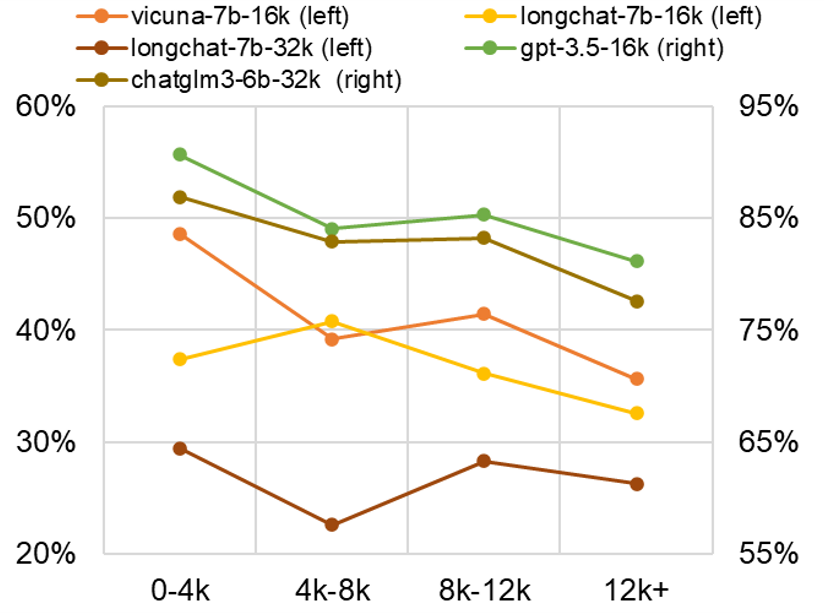}}
\subfigure[Input Position]{
    \label{fig:lost_in_the_middle}
    \includegraphics[scale=0.38]{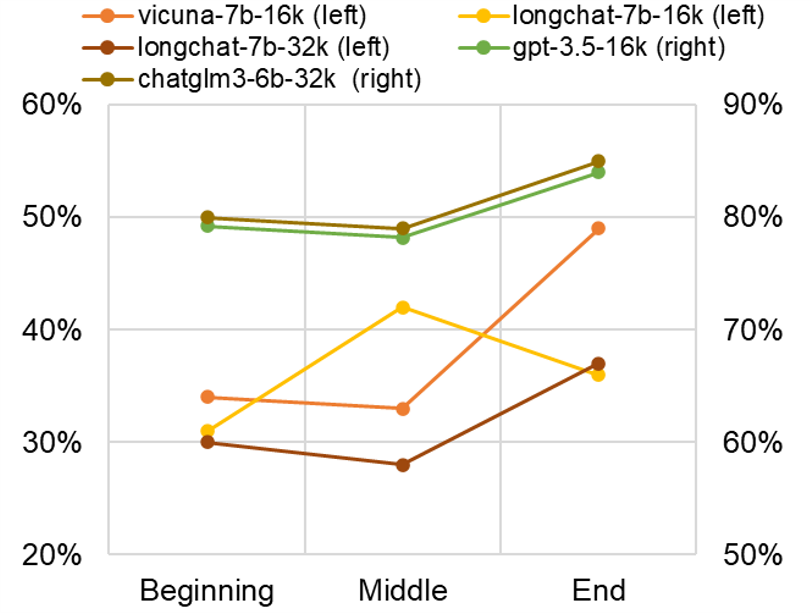}}
\subfigure[Acc\_by\_Dates]{
    \label{fig:experi_retri_order_accbydate}
    \includegraphics[scale=0.38]{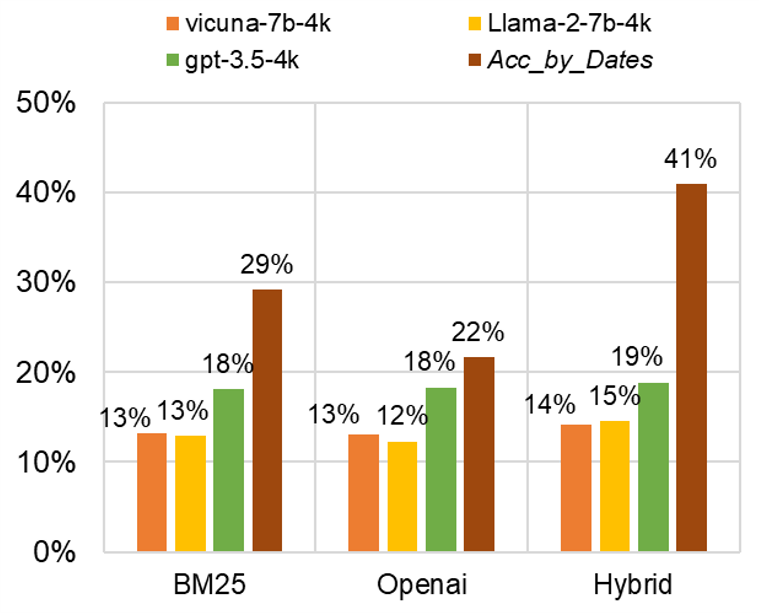}}
    \vspace{-0.2cm}
\caption{Analysis of results on TCELongBench. (a) shows the average accuracy under different context length in TLB-detail; (b) demonstrates "Lost in the middle" phenomenon in TLB-detail, except for LongChat-16k; (c) shows the $Acc\_by\_Dates$ scores under Retrieve One by One strategy in TLB-order.}
\vspace{-0.3cm}
\label{level}
\end{figure*}

\subsection{Main Results}

The results are reported in Table \ref{table:all_exp}. It is clear and as expected that \texttt{gpt-4-128k} outperforms all other models by a significant margin for all close-ended questions. Lower accuracy scores of MCQs in TLB-forecast than TLB-detail indicates forecasting future event is a more challenging task. Moreover, all models perform poorly in the open-domain of TLB-forecast, where context only brings slight improvement. Additionally, increasing model size drives the performance of Vicuna and Llama-2 upwards across all tasks.

\textbf{Retriever emerges as a performance bottleneck for models leveraging RAG method.}
Results of retrievers' performance in Table \ref{table:retri_exp} offer insights into the varying performance of the same model with different retrievers, as illustrated in Table \ref{table:all_exp}. Specifically, hybrid retriever demonstrates the most optimal performance for each model in TLB-detail, while BM25 and Hybrid retrievers brings out better performance in TLB-order under two strategies respectively.

\textbf{Retrievers may not consistently yield effective results.} When concatenating three choices in the ranking problem for retrieval, i.e. strategy-1 discussed in Section \ref{sc:analysis}, retrievers yield slightly improved performance for open-source 7B models, but worsened performance for open-source 13B models and the close-source model. This observation suggests that inappropriate context can be misleading, particularly for more powerful models. Such discrepancies may arise from potential data leakage during their training stages.

\textbf{Long context modeling techniques offer benefits for temporal sequencing, but may lead to inferior performance.}
\texttt{gpt-3.5-16k} and \texttt{chatglm3-6b-32k} achieve comparable performance with \texttt{gpt-3.5-4k} with hybrid retriever, and even perform better in TLB-order. 
However, \texttt{vicuna-7b-16k}, \texttt{longchat-7b-16k} and \texttt{longchat-7b-32k} underperform retrieval-augmented models by a significant margin. This finding indicates that fintuning longer is still challenging and may lead to inferior performance, while its upper limit could achieve even better performance than RAG method.

\begin{table}[h!]
\setlength\tabcolsep{2pt}
\centering
\renewcommand{\arraystretch}{1.0}
\scalebox{0.75}{
\begin{tabular}{l c c c c} 
 \specialrule{0.08em}{3pt}{3pt}
 \multirow{2}*{\textbf{Retriever}} & \multicolumn{2}{c}{\textbf{TLB-detail}} & \multicolumn{2}{c}{\textbf{TLB-order}} \\
  & $Acc\_Doc$ & $Acc\_Date$ & $Acc\_Dates$-1 & $Acc\_Dates$-2  \\
 \specialrule{0.04em}{3pt}{3pt}
 BM25 & 72.8 & 85.1  & \textbf{15.7} & 16.2   \\
 Openai & 64.9 & 79.1  & 5.9 & 10.9  \\ 
 Hybrid & \textbf{75.3} & \textbf{87.5}  & 1.1 & \textbf{26.7} \\
 \specialrule{0.08em}{3pt}{3pt}
\end{tabular}}
\vspace{-0.2cm}
\caption{Performance of retrievers, where "-1" and "-2" indicate \textit{Retrieving Once} strategy and \textit{Retrieving One by One} strategy respectively.}
\vspace{-0.3cm}
\label{table:retri_exp}
\end{table}

\subsection{Detailed Analysis} 
\label{sc:analysis}

We conduct detailed analysis on the experiment results of TCELongBench from various aspects.

\noindent{\textbf{Impact of Input Length and Position.}} 
For fine-grained analysis of context of models with long context window, we explore how their performance in TLB-detail varies across different context length ranges of 0-4k, 4k-8k, 8k-12k, and 12k+ \footnote{QA pairs are divided into various ranges by tokenizing their contexts using vicuna-16k and counting token numbers.}. The slopes of curves in Figure \ref{fig:distr_long_qa_acc} showcase a drop in performance on data of greater length. 

Furthermore, we investigate the impact of the position of relevant articles on the model's performance \cite{liu2023lost} in TLB-detail. In specific, we experiment with relocating articles with gold timestamps to different positions within the context window, using a random sample size of 100. As shown in Figure \ref{fig:lost_in_the_middle}, most LLMs exhibit improved accuracy towards the end, for questions also being situated at the end of the prompt (see Table \ref{table:prompt_eval_detail}), except for \texttt{longchat-7b-16k}.

\noindent{\textbf{Retrieving for Temporal Sequencing.}}
We employ two retrieving strategies in TLB-order: (1) \textit{Retrieve Once} strategy concatenates three choices together to retrieve top three text chunks; (2) \textit{Retrieve One by One} strategy retrieves each choice and then select the text chunk with the earliest timestamp from the top three -- the news articles often repeat the reports in earlier days. 

Strategy-2 consistently leads to model's better performance than strategy-1, as shown in Table \ref{table:all_exp}. This finding is explained by results reported in Table \ref{table:retri_exp}, where retrievers achieve higher $Acc\_Dates$ scores in strategy-2. Moreover, the combination of hybrid retriever and strategy-2 demonstrates the most optimal performance among most models.

Additionally, candidate choices in strategy-2 could be directly ranked according to the timestamps of retrieved text chunks, that is, no LLMs involved. This accuracy score is labeled as $Acc\_by\_Date$ in Figure \ref{fig:experi_retri_order_accbydate}, where we can see that this straightforward approach outperforms others by a considerable margin. This finding demonstrates that LLMs hardly leverage the full temporal information via ICL, even though all timestamps are fed into LLMs with clear format. Incorporating further time-aware instruction tuning could be beneficial, a direction we consider for future research.

\noindent{\textbf{Open-Domain Error Analysis.}} We observe that LLMs tend to give lengthy and indirect answers to forecasting questions by using expressions like "It is not possible to accurately forecast what", and "It is difficult to say with 100\%". Inspired by \citet{kamalloo-etal-2023-evaluating}, we classify a sample of these open answers into three categories: \textit{Semantically Correct}, \textit{Wrong}, and \textit{Invalid}. Specifically, \textit{Semantically Correct} answer is semantically equivalent to the ground truth, while \textit{Invalid} answer suggests that the model refuses to give a clear answer to the forecasting question. 

\begin{figure}[htp]
    \centering
    \includegraphics[scale=0.5]{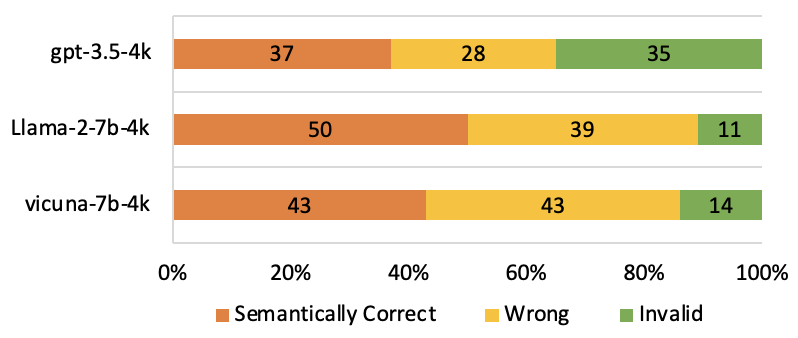}
    \caption{Classification of open-domain answers to 100 random questions in TLB-forecast. The three 4k models are with hybrid retriever.}
    \label{fig:Forecast_error_analysis}
\end{figure}

We randomly sample 100 forecasting questions and collect their corresponding of by each 4k model with hybrid retriever. As shown in Figure  \ref{fig:Forecast_error_analysis}, \texttt{Llama-2-7b-4k} outputs more semantically correct answers than \texttt{vicuna-7b-4k} within the random sample, inconsistent with results in Table \ref{table:all_exp}. \texttt{gpt-3.5-4k} gives the most invalid answers, probably due to stringent safety-alignment technique. 

\section{Conclusion}

In this work, we introduced a LLM-based framework for outline extraction of TCE and established TCELongBench to evaluate LLMs' capability of temporal understanding and long text comprehension. Our approach involved three tasks targeting reading comprehension, temporal sequencing, and future event forecasting, and conducted experiments across two foundational models: LLMs leveraging RAG method and LLMs with long context windows. While our experiments provided valuable insights into LLMs' abilities in TCE analysis, future research is essential, particularly in content generation tasks \cite{reddy2023smartbook}, to unlock the full potential of LLMs in complex narrative understanding.

\section*{Limitation}
Our work focuses on evaluating LLM's capability of temporal, long text understanding using test sets of TCELongBench. Thus, we do not utilize the training and development sets, reserving them for future work. 

We do not differentiate whether or not news articles in TCELongBench are included in the massive training data of LLMs. This explains why \texttt{gpt-3.5-4k} achieves over 50\% accuracy of MCQs without any context -- some news articles may be already memorized by LLMs during training stage. Nonetheless, our dataset construction pipeline is adaptable to new, unseen corpora, which will be the focus of our future research. 

During experiments, we design prompt templates to instruct LLMs to output their answers under some specific formats (see Appendix \ref{ap:prompt_strategy_all}). Answers that do not follow these formats would be regarded as incorrect answers, which leads to the loss of model's performance. Additionally, some parameters in the experiment setting, such as the number and size of retrieved chunks, could be further adjusted to discover new insights. Due to the content length and time limitation, we set these parameters to fixed values.

\bibliography{anthology}

\begin{thebibliography}{36}
\expandafter\ifx\csname natexlab\endcsname\relax\def\natexlab#1{#1}\fi

\bibitem[{An et~al.(2023)An, Gong, Zhong, Zhao, Li, Zhang, Kong, and Qiu}]{an2023leval}
Chenxin An, Shansan Gong, Ming Zhong, Xingjian Zhao, Mukai Li, Jun Zhang, Lingpeng Kong, and Xipeng Qiu. 2023.
\newblock \href {http://arxiv.org/abs/2307.11088} {L-eval: Instituting standardized evaluation for long context language models}.

\bibitem[{Bai et~al.(2023)Bai, Lv, Zhang, Lyu, Tang, Huang, Du, Liu, Zeng, Hou, Dong, Tang, and Li}]{bai2023longbench}
Yushi Bai, Xin Lv, Jiajie Zhang, Hongchang Lyu, Jiankai Tang, Zhidian Huang, Zhengxiao Du, Xiao Liu, Aohan Zeng, Lei Hou, Yuxiao Dong, Jie Tang, and Juanzi Li. 2023.
\newblock \href {http://arxiv.org/abs/2308.14508} {Longbench: A bilingual, multitask benchmark for long context understanding}.

\bibitem[{Banerjee and Lavie(2005)}]{banerjee-lavie-2005-meteor}
Satanjeev Banerjee and Alon Lavie. 2005.
\newblock \href {https://aclanthology.org/W05-0909} {{METEOR}: An automatic metric for {MT} evaluation with improved correlation with human judgments}.
\newblock In \emph{Proceedings of the {ACL} Workshop on Intrinsic and Extrinsic Evaluation Measures for Machine Translation and/or Summarization}, pages 65--72. Association for Computational Linguistics.

\bibitem[{Berzak et~al.(2020)Berzak, Malmaud, and Levy}]{berzak-etal-2020-starc}
Yevgeni Berzak, Jonathan Malmaud, and Roger Levy. 2020.
\newblock \href {https://doi.org/10.18653/v1/2020.acl-main.507} {{STARC}: Structured annotations for reading comprehension}.
\newblock In \emph{Proceedings of the 58th Annual Meeting of the Association for Computational Linguistics}, pages 5726--5735. Association for Computational Linguistics.

\bibitem[{Brown et~al.(2020)Brown, Mann, Ryder, Subbiah, Kaplan, Dhariwal, Neelakantan, Shyam, Sastry, Askell, Agarwal, Herbert-Voss, Krueger, Henighan, Child, Ramesh, Ziegler, Wu, Winter, Hesse, Chen, Sigler, Litwin, Gray, Chess, Clark, Berner, McCandlish, Radford, Sutskever, and Amodei}]{NEURIPS2020_1457c0d6}
Tom Brown, Benjamin Mann, Nick Ryder, Melanie Subbiah, Jared~D Kaplan, Prafulla Dhariwal, Arvind Neelakantan, Pranav Shyam, Girish Sastry, Amanda Askell, Sandhini Agarwal, Ariel Herbert-Voss, Gretchen Krueger, Tom Henighan, Rewon Child, Aditya Ramesh, Daniel Ziegler, Jeffrey Wu, Clemens Winter, Chris Hesse, Mark Chen, Eric Sigler, Mateusz Litwin, Scott Gray, Benjamin Chess, Jack Clark, Christopher Berner, Sam McCandlish, Alec Radford, Ilya Sutskever, and Dario Amodei. 2020.
\newblock \href {https://proceedings.neurips.cc/paper_files/paper/2020/file/1457c0d6bfcb4967418bfb8ac142f64a-Paper.pdf} {Language models are few-shot learners}.
\newblock In \emph{Advances in Neural Information Processing Systems}, volume~33, pages 1877--1901. Curran Associates, Inc.

\bibitem[{Dhingra et~al.(2022)Dhingra, Cole, Eisenschlos, Gillick, Eisenstein, and Cohen}]{dhingra-etal-2022-time}
Bhuwan Dhingra, Jeremy~R. Cole, Julian~Martin Eisenschlos, Daniel Gillick, Jacob Eisenstein, and William~W. Cohen. 2022.
\newblock \href {https://doi.org/10.1162/tacl_a_00459} {Time-aware language models as temporal knowledge bases}.
\newblock \emph{Transactions of the Association for Computational Linguistics}, 10:257--273.

\bibitem[{Dong et~al.(2023)Dong, Tang, Li, Zhao, and Wen}]{dong2023bamboo}
Zican Dong, Tianyi Tang, Junyi Li, Wayne~Xin Zhao, and Ji-Rong Wen. 2023.
\newblock \href {http://arxiv.org/abs/2309.13345} {Bamboo: A comprehensive benchmark for evaluating long text modeling capacities of large language models}.

\bibitem[{Du et~al.(2022)Du, Zhang, Li, Yu, Wang, Lai, Lin, Wang, Liu, Zhou, Wen, Li, Hannan, Lei, Kim, Dror, Wang, Regan, Zeng, Lyu, Yu, Edwards, Jin, Jiao, Kazeminejad, Wang, Callison-Burch, Bansal, Vondrick, Han, Roth, Chang, Palmer, and Ji}]{du-etal-2022-resin}
Xinya Du, Zixuan Zhang, Sha Li, Pengfei Yu, Hongwei Wang, Tuan Lai, Xudong Lin, Ziqi Wang, Iris Liu, Ben Zhou, Haoyang Wen, Manling Li, Darryl Hannan, Jie Lei, Hyounghun Kim, Rotem Dror, Haoyu Wang, Michael Regan, Qi~Zeng, Qing Lyu, Charles Yu, Carl Edwards, Xiaomeng Jin, Yizhu Jiao, Ghazaleh Kazeminejad, Zhenhailong Wang, Chris Callison-Burch, Mohit Bansal, Carl Vondrick, Jiawei Han, Dan Roth, Shih-Fu Chang, Martha Palmer, and Heng Ji. 2022.
\newblock \href {https://doi.org/10.18653/v1/2022.naacl-demo.7} {{RESIN}-11: Schema-guided event prediction for 11 newsworthy scenarios}.
\newblock In \emph{Proceedings of the 2022 Conference of the North American Chapter of the Association for Computational Linguistics: Human Language Technologies: System Demonstrations}, pages 54--63. Association for Computational Linguistics.

\bibitem[{Gao et~al.(2021)Gao, Yao, and Chen}]{gao2021simcse}
Tianyu Gao, Xingcheng Yao, and Danqi Chen. 2021.
\newblock \href {https://doi.org/10.18653/v1/2021.emnlp-main.552} {{S}im{CSE}: Simple contrastive learning of sentence embeddings}.
\newblock In \emph{Proceedings of the 2021 Conference on Empirical Methods in Natural Language Processing}, pages 6894--6910. Association for Computational Linguistics.

\bibitem[{Gholipour~Ghalandari et~al.(2020)Gholipour~Ghalandari, Hokamp, Pham, Glover, and Ifrim}]{gholipour-ghalandari-etal-2020-large}
Demian Gholipour~Ghalandari, Chris Hokamp, Nghia~The Pham, John Glover, and Georgiana Ifrim. 2020.
\newblock \href {https://doi.org/10.18653/v1/2020.acl-main.120} {A large-scale multi-document summarization dataset from the {W}ikipedia current events portal}.
\newblock In \emph{Proceedings of the 58th Annual Meeting of the Association for Computational Linguistics}, pages 1302--1308. Association for Computational Linguistics.

\bibitem[{Gholipour~Ghalandari and Ifrim(2020)}]{gholipour-ghalandari-ifrim-2020-examining}
Demian Gholipour~Ghalandari and Georgiana Ifrim. 2020.
\newblock \href {https://doi.org/10.18653/v1/2020.acl-main.122} {Examining the state-of-the-art in news timeline summarization}.
\newblock In \emph{Proceedings of the 58th Annual Meeting of the Association for Computational Linguistics}, pages 1322--1334, Online. Association for Computational Linguistics.

\bibitem[{Haladyna et~al.(2002)Haladyna, Downing, and Rodriguez}]{itemwriteflaws}
Thomas Haladyna, Steven Downing, and Michael Rodriguez. 2002.
\newblock \href {https://doi.org/10.1207/S15324818AME1503_5} {A review of multiple-choice item-writing guidelines for classroom assessment}.
\newblock \emph{Applied Measurement in Education - APPL MEAS EDUC}, 15:309--333.

\bibitem[{Honnibal and Montani(2017)}]{spacy2}
Matthew Honnibal and Ines Montani. 2017.
\newblock {spaCy 2}: Natural language understanding with {B}loom embeddings, convolutional neural networks and incremental parsing.
\newblock To appear.

\bibitem[{Jiao et~al.(2023)Jiao, Zhong, Shen, Zhang, Zhang, and Han}]{10.1145/3543507.3583295}
Yizhu Jiao, Ming Zhong, Jiaming Shen, Yunyi Zhang, Chao Zhang, and Jiawei Han. 2023.
\newblock \href {https://doi.org/10.1145/3543507.3583295} {Unsupervised event chain mining from multiple documents}.
\newblock In \emph{Proceedings of the ACM Web Conference 2023}, WWW '23, page 1948–1959. Association for Computing Machinery.

\bibitem[{Jin et~al.(2021)Jin, Khanna, Kim, Lee, Morstatter, Galstyan, and Ren}]{jin-etal-2021-forecastqa}
Woojeong Jin, Rahul Khanna, Suji Kim, Dong-Ho Lee, Fred Morstatter, Aram Galstyan, and Xiang Ren. 2021.
\newblock \href {https://doi.org/10.18653/v1/2021.acl-long.357} {{F}orecast{QA}: A question answering challenge for event forecasting with temporal text data}.
\newblock In \emph{Proceedings of the 59th Annual Meeting of the Association for Computational Linguistics and the 11th International Joint Conference on Natural Language Processing (Volume 1: Long Papers)}, pages 4636--4650. Association for Computational Linguistics.

\bibitem[{Kamalloo et~al.(2023)Kamalloo, Dziri, Clarke, and Rafiei}]{kamalloo-etal-2023-evaluating}
Ehsan Kamalloo, Nouha Dziri, Charles Clarke, and Davood Rafiei. 2023.
\newblock \href {https://doi.org/10.18653/v1/2023.acl-long.307} {Evaluating open-domain question answering in the era of large language models}.
\newblock In \emph{Proceedings of the 61st Annual Meeting of the Association for Computational Linguistics (Volume 1: Long Papers)}, pages 5591--5606. Association for Computational Linguistics.

\bibitem[{Li et~al.(2021)Li, Li, Wang, Huang, Cho, Ji, Han, and Voss}]{li-etal-2021-future}
Manling Li, Sha Li, Zhenhailong Wang, Lifu Huang, Kyunghyun Cho, Heng Ji, Jiawei Han, and Clare Voss. 2021.
\newblock \href {https://doi.org/10.18653/v1/2021.emnlp-main.422} {The future is not one-dimensional: Complex event schema induction by graph modeling for event prediction}.
\newblock In \emph{Proceedings of the 2021 Conference on Empirical Methods in Natural Language Processing}, pages 5203--5215. Association for Computational Linguistics.

\bibitem[{Lin et~al.(2021)Lin, Chambers, and Durrett}]{lin-etal-2021-conditional}
Shih-Ting Lin, Nathanael Chambers, and Greg Durrett. 2021.
\newblock \href {https://doi.org/10.18653/v1/2021.acl-long.555} {Conditional generation of temporally-ordered event sequences}.
\newblock In \emph{Proceedings of the 59th Annual Meeting of the Association for Computational Linguistics and the 11th International Joint Conference on Natural Language Processing (Volume 1: Long Papers)}, pages 7142--7157. Association for Computational Linguistics.

\bibitem[{Liu(2022)}]{Liu_LlamaIndex_2022}
Jerry Liu. 2022.
\newblock \href {https://doi.org/10.5281/zenodo.1234} {{LlamaIndex}}.

\bibitem[{Liu et~al.(2023)Liu, Lin, Hewitt, Paranjape, Bevilacqua, Petroni, and Liang}]{liu2023lost}
Nelson~F. Liu, Kevin Lin, John Hewitt, Ashwin Paranjape, Michele Bevilacqua, Fabio Petroni, and Percy Liang. 2023.
\newblock \href {http://arxiv.org/abs/2307.03172} {Lost in the middle: How language models use long contexts}.

\bibitem[{Ma et~al.(2023)Ma, Ye, Wu, Wang, Cao, Pang, and Chua}]{ma2023structured}
Yunshan Ma, Chenchen Ye, Zijian Wu, Xiang Wang, Yixin Cao, Liang Pang, and Tat-Seng Chua. 2023.
\newblock \href {http://arxiv.org/abs/2312.01052} {Structured, complex and time-complete temporal event forecasting}.

\bibitem[{Miller et~al.(2009)Miller, Vandome, and McBrewster}]{lev_dist}
Frederic~P. Miller, Agnes~F. Vandome, and John McBrewster. 2009.
\newblock \emph{Levenshtein Distance: Information Theory, Computer Science, String (Computer Science), String Metric, Damerau?Levenshtein Distance, Spell Checker, Hamming Distance}.
\newblock Alpha Press.

\bibitem[{Nijkamp et~al.(2023)Nijkamp, Xie, Hayashi, Pang, Xia, Xing, Vig, Yavuz, Laban, Krause, Purushwalkam, Niu, Kryściński, Murakhovs'ka, Choubey, Fabbri, Liu, Meng, Tu, Bhat, Wu, Savarese, Zhou, Joty, and Xiong}]{nijkamp2023xgen7b}
Erik Nijkamp, Tian Xie, Hiroaki Hayashi, Bo~Pang, Congying Xia, Chen Xing, Jesse Vig, Semih Yavuz, Philippe Laban, Ben Krause, Senthil Purushwalkam, Tong Niu, Wojciech Kryściński, Lidiya Murakhovs'ka, Prafulla~Kumar Choubey, Alex Fabbri, Ye~Liu, Rui Meng, Lifu Tu, Meghana Bhat, Chien-Sheng Wu, Silvio Savarese, Yingbo Zhou, Shafiq Joty, and Caiming Xiong. 2023.
\newblock \href {http://arxiv.org/abs/2309.03450} {Xgen-7b technical report}.

\bibitem[{Papineni et~al.(2002)Papineni, Roukos, Ward, and Zhu}]{10.3115/1073083.1073135}
Kishore Papineni, Salim Roukos, Todd Ward, and Wei-Jing Zhu. 2002.
\newblock \href {https://doi.org/10.3115/1073083.1073135} {Bleu: a method for automatic evaluation of machine translation}.
\newblock In \emph{Proceedings of the 40th Annual Meeting on Association for Computational Linguistics}, ACL '02, page 311–318. Association for Computational Linguistics.

\bibitem[{Rashkin et~al.(2020)Rashkin, Celikyilmaz, Choi, and Gao}]{rashkin-etal-2020-plotmachines}
Hannah Rashkin, Asli Celikyilmaz, Yejin Choi, and Jianfeng Gao. 2020.
\newblock \href {https://doi.org/10.18653/v1/2020.emnlp-main.349} {{P}lot{M}achines: Outline-conditioned generation with dynamic plot state tracking}.
\newblock In \emph{Proceedings of the 2020 Conference on Empirical Methods in Natural Language Processing (EMNLP)}, pages 4274--4295. Association for Computational Linguistics.

\bibitem[{Reddy et~al.(2023)Reddy, Fung, Zeng, Li, Wang, Sullivan, and Ji}]{reddy2023smartbook}
Revanth~Gangi Reddy, Yi~R. Fung, Qi~Zeng, Manling Li, Ziqi Wang, Paul Sullivan, and Heng Ji. 2023.
\newblock \href {http://arxiv.org/abs/2303.14337} {Smartbook: Ai-assisted situation report generation}.

\bibitem[{Reimers and Gurevych(2020)}]{reimers-2020-multilingual-sentence-bert}
Nils Reimers and Iryna Gurevych. 2020.
\newblock \href {https://arxiv.org/abs/2004.09813} {Making monolingual sentence embeddings multilingual using knowledge distillation}.
\newblock In \emph{Proceedings of the 2020 Conference on Empirical Methods in Natural Language Processing}. Association for Computational Linguistics.

\bibitem[{Saravanakumar et~al.(2021)Saravanakumar, Ballesteros, Chandrasekaran, and McKeown}]{saravanakumar-etal-2021-event}
Kailash~Karthik Saravanakumar, Miguel Ballesteros, Muthu~Kumar Chandrasekaran, and Kathleen McKeown. 2021.
\newblock \href {https://doi.org/10.18653/v1/2021.eacl-main.198} {Event-driven news stream clustering using entity-aware contextual embeddings}.
\newblock In \emph{Proceedings of the 16th Conference of the European Chapter of the Association for Computational Linguistics: Main Volume}, pages 2330--2340. Association for Computational Linguistics.

\bibitem[{Shaham et~al.(2023)Shaham, Ivgi, Efrat, Berant, and Levy}]{shaham2023zeroscrolls}
Uri Shaham, Maor Ivgi, Avia Efrat, Jonathan Berant, and Omer Levy. 2023.
\newblock \href {https://doi.org/10.18653/v1/2023.findings-emnlp.536} {{Z}ero{SCROLLS}: A zero-shot benchmark for long text understanding}.
\newblock In \emph{Findings of the Association for Computational Linguistics: EMNLP 2023}, pages 7977--7989. Association for Computational Linguistics.

\bibitem[{Steen and Markert(2019)}]{steen-markert-2019-abstractive}
Julius Steen and Katja Markert. 2019.
\newblock \href {https://doi.org/10.18653/v1/D19-5403} {Abstractive timeline summarization}.
\newblock In \emph{Proceedings of the 2nd Workshop on New Frontiers in Summarization}, pages 21--31, Hong Kong, China. Association for Computational Linguistics.

\bibitem[{Tan et~al.(2023)Tan, Ng, and Bing}]{tan2023benchmarking}
Qingyu Tan, Hwee~Tou Ng, and Lidong Bing. 2023.
\newblock \href {https://doi.org/10.18653/v1/2023.acl-long.828} {Towards benchmarking and improving the temporal reasoning capability of large language models}.
\newblock In \emph{Proceedings of the 61st Annual Meeting of the Association for Computational Linguistics (Volume 1: Long Papers)}, pages 14820--14835. Association for Computational Linguistics.

\bibitem[{Wang and Zhao(2023)}]{wang2023tram}
Yuqing Wang and Yun Zhao. 2023.
\newblock \href {http://arxiv.org/abs/2310.00835} {Tram: Benchmarking temporal reasoning for large language models}.

\bibitem[{Xu et~al.(2024)Xu, Ping, Wu, McAfee, Zhu, Liu, Subramanian, Bakhturina, Shoeybi, and Catanzaro}]{xu2024retrieval}
Peng Xu, Wei Ping, Xianchao Wu, Lawrence McAfee, Chen Zhu, Zihan Liu, Sandeep Subramanian, Evelina Bakhturina, Mohammad Shoeybi, and Bryan Catanzaro. 2024.
\newblock \href {https://openreview.net/forum?id=xw5nxFWMlo} {Retrieval meets long context large language models}.
\newblock In \emph{The Twelfth International Conference on Learning Representations}.

\bibitem[{Yoon et~al.(2023)Yoon, Meng, Lee, and Han}]{10.1145/3543507.3583507}
Susik Yoon, Yu~Meng, Dongha Lee, and Jiawei Han. 2023.
\newblock \href {https://doi.org/10.1145/3543507.3583507} {Scstory: Self-supervised and continual online story discovery}.
\newblock In \emph{Proceedings of the ACM Web Conference 2023}, WWW '23, page 1853–1864. Association for Computing Machinery.

\bibitem[{Zhang and Choi(2021)}]{zhang-choi-2021-situatedqa}
Michael Zhang and Eunsol Choi. 2021.
\newblock \href {https://doi.org/10.18653/v1/2021.emnlp-main.586} {{S}ituated{QA}: Incorporating extra-linguistic contexts into {QA}}.
\newblock In \emph{Proceedings of the 2021 Conference on Empirical Methods in Natural Language Processing}, pages 7371--7387. Association for Computational Linguistics.

\bibitem[{Zhu et~al.(2023)Zhu, Zhang, Gao, Qin, Xu, and Yang}]{zhu-etal-2023-diffusion}
Fangqi Zhu, Lin Zhang, Jun Gao, Bing Qin, Ruifeng Xu, and Haiqin Yang. 2023.
\newblock \href {https://doi.org/10.18653/v1/2023.findings-acl.800} {A diffusion model for event skeleton generation}.
\newblock In \emph{Findings of the Association for Computational Linguistics: ACL 2023}, pages 12630--12641. Association for Computational Linguistics.

\end{thebibliography}

\newpage
\appendix

\section{Dataset}
% \label{ap:appendix}

\subsection{Deduplication}
\label{ap:point_dupli}
We conduct multiple deduplication procedures throughout outline extraction and dataset construction. This is conducted by calculating two similarity scores using \texttt{sup-simcse-bert} \footnote{https://huggingface.co/princeton-nlp/sup-simcse-bert-base-uncased} \cite{gao2021simcse} and \texttt{quora-distilroberta} \footnote{https://huggingface.co/cross-encoder/quora-distilroberta-base} \cite{reimers-2020-multilingual-sentence-bert}. While \texttt{quora-distilroberta} is specialized in detecting duplicated questions, \texttt{sup-simcse-bert} offers high-quality sentence embeddings to decide whether two sentences are semantically equivalent based on the similarity score of their embeddings. Both thresholds are set to 0.8 based on our observations in practice. Note that QA pairs in TLB-order is deduplicated by the common key points instead of similarity scores.

The proportion of discarded key points and QA pairs in TCELongBench are shown in Table \ref{table:sim_dup}. Note that we also discard the noising key points if their similarity scores are below 0.2 with others in the same TCE, since they may be the regular greetings of LLMs, incomplete sentences, etc.

\begin{table}[h!]
\centering
\scalebox{0.85}{
\begin{tabular}{l r r c } 
 \hline
  & \textbf{Before} & \textbf{After} & \textbf{\%}  \\
 \hline
 Key Point & 137,041 & 91,574 & 33.2 \\
 TLB-detail & 74,568 & 61,053 & 18.2 \\
 TLB-order & 55,663 & 21,164 & 62.0 \\
 TLB-forecast & 7,664 & 6,604 &  13.8 \\
 \hline
\end{tabular}}
\caption{Numbers of key points and QA pairs in TCELongBench before and after de-duplication, and the proportions of de-duplicated ones.}
\label{table:sim_dup}
\end{table}

\subsection{Human Evaluation}
\label{ap:hu_eval}

% [htp]
% [h!]
\begin{table*}[htp]
\small
\centering
\begin{tabular}{l | c | c c c c c c c} 
 \specialrule{0.08em}{3pt}{3pt}
 \textbf{Dataset} & \textbf{Num} & \textbf{Acc.} & \textbf{Context} & \textbf{Reasonable} & \textbf{Plausible} & \textbf{Temporal} & \textbf{Storytelling} & \textbf{Evidence(\&Unseen)} \\
 \specialrule{0.08em}{3pt}{3pt}
 TLB-detail & 30 & 85.56  & 95.56 & 95.56 & 84.44 &  &  & 94.44  \\
 TLB-order & 30 & 71.11  & 98.89 &  &  & 77.78 & 95.56 & 86.67 \\
 TLB-forecast & 24 & 75.00  & 98.61 & 95.83 & 97.22 &  &  & 77.78 \\
 \specialrule{0.04em}{2pt}{2pt}
 \textit{Total} & 84 & 77.38  & 97.61 & 95.67 & 90.12 & 77.78 & 95.56 & 86.90 \\
 \specialrule{0.08em}{3pt}{3pt}
\end{tabular}
\caption{Results of Human Evaluation by three annotators. The unit of all figures are percent \% except Num.}
\label{ap_table:hu_eval}
\end{table*}

We evaluate the quality of our QA datasets from multiple dimensions. For TLB-detail, we evaluate from five dimensions below:
\begin{itemize}[leftmargin=*, itemsep=1pt, topsep=1pt, parsep=1pt]
    \item \textit{Human Performance}. Annotators are asked to answer multiple choice questions with access to all documents except those on the last days of complex events, and record their accuracy scores.
    \item \textit{Context}. We want to see whether the annotators need the context from the documents to understand and answer the question with confidence.
    \item Evidence. This is to check whether the annotators are able to find the evidence from the documents to support the correct answer.
    \item \textit{Reasonable}. Inspired by \cite{itemwriteflaws}, \textit{Reasonable} evaluates the quality of question from three aspects, namely clear, clueless and focused. A clear, clueless and focused question is written in clear and unambiguous language, brings no grammatical or logical cue to the correct answer, and does not contain unnecessary information that is not required to answer it.
    \item \textit{Plausible}. Inspired by \cite{itemwriteflaws}, \textit{Plausible} evaluates the quality of four choices from two aspects, namely similar and unique. While all four choices are plausible to the question and homogeneous in wording, they should be essentially different so that there is only one correct answer.
\end{itemize}

For TLB-forecast, we inherit all five dimensions from TLB-detail, and modify \textit{Evidence} to \textit{Correct\&Unseen}. \textit{Evidence\&Unseen} does not only require finding the supporting evidence from the articles on the last day, but also check if the annotators are unable to answer the question with 100\% certainty given the articles in former days.

For TLB-order, we inherit three dimensions from TLB-detail, \textit{Human Performance}, \textit{Context}, and \textit{Evidence}, and add two new dimensions \textit{Temporal} and \textit{Storytelling} shown below. Note that \textit{Evidence} here is to check if each of the choice indeed comes from the documents in its timestamps, since it is likely that the choice's content may already exist in the earlier timestamp for summarizing documents in each day sacrificing many details.
\begin{itemize}[leftmargin=*, itemsep=1pt, topsep=1pt, parsep=1pt]
    \item \textit{Temporal}. This dimension requires the choice's content presenting the event that just happened or was happening, instead of the event that had happened over a time or may happen in the future.
    \item \textit{Storytelling}. We ask the annotators to check whether the choices in the correct order present a brief storyline with potential logic and are connected by common entities.
\end{itemize}

We give the detailed definitions of above dimensions, as instructions, to three annotators for human evaluation. They are postgraduate students from China and Singapore, proficient in English reading. Detailed results of human evaluation is shown in Table \ref{ap_table:hu_eval}. Most QA pairs satisfy the requirements of all dimensions. 

\subsection{Quality of Choices in MCQ}
\begin{figure}[thp]
\centering
\vspace{-0.35cm}
\subfigtopskip=2pt
\subfigbottomskip=2pt
\subfigcapskip=-5pt
\subfigure[TLB-detail]{
    \label{fig:four_choice_distr_MCQ}
    \includegraphics[scale=0.35]{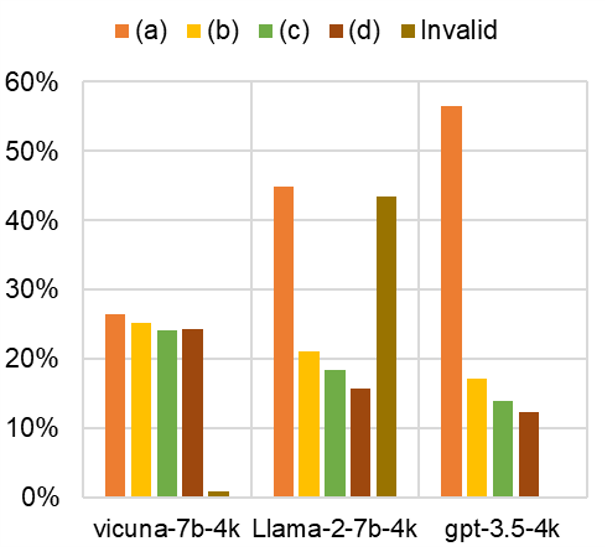}}
\subfigure[TLB-forecast]{
    \label{fig:four_choice_distr_fore}
    \includegraphics[scale=0.35]{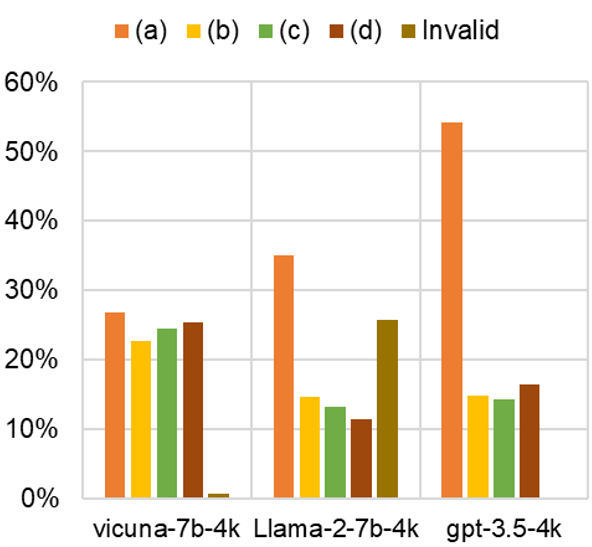}}
\caption{Distribution of four choices of experiment results of (a) TLB-detail and (b) TLB-forecast when without any context.}
\label{level}
\end{figure}

To further check the quality of misleading answers, we calculate the proportions of four choices selected by LLMs during evaluating without any context. Recall that (a) is the correct answer while (b), (c) and (d) are misleading answers. As shown in Figure \ref{fig:four_choice_distr_fore}, \texttt{vicuna-7b-4k} select four candidate choices with nearly equal probability, proving the high-quality of our misleading answers, while \texttt{Llama-2-7b-4k} generate the most invalid answers that do not follow the output format. \texttt{gpt-3.5-4k} achieve over 50\% accuracy scores without any context, due to the data leakage during training stage.

\section{Experiment}

\subsection{Baseline Models}

For LLM with 4k context window, models in our experiments are listed below:
\begin{itemize}[leftmargin=*, itemsep=1pt, topsep=1pt, parsep=1pt]
    \item \texttt{vicuna-7b-4k} \footnote{https://huggingface.co/lmsys/vicuna-7b-v1.5} and \texttt{vicuna-13b-4k} \footnote{https://huggingface.co/lmsys/vicuna-13b-v1.5} are both Vicuna v1.5, fine-tuned from Llama 2 with supervised instruction fine-tuning.
    \item \texttt{Llama-2-7b-4k} \footnote{https://huggingface.co/meta-llama/Llama-2-7b-chat-hf} and \texttt{Llama-2-13b-4k} \footnote{https://huggingface.co/meta-llama/Llama-2-13b-chat-hf} are chatbots based on Llama 2 released by Meta AI.
    \item \texttt{gpt-3.5-4k} \footnote{https://platform.openai.com/docs/models/gpt-3-5-turbo} is gpt-3.5-turbo-0613 model provided by OpenAI.
\end{itemize}

For LLM with long context window, models in our experiments are listed below:
\begin{itemize}[leftmargin=*, itemsep=1pt, topsep=1pt, parsep=1pt]
    \item \texttt{vicuna-7b-16k} \footnote{https://huggingface.co/lmsys/vicuna-7b-v1.5-16k} is Vicuna v1.5, fine-tuned from Llama 2 with supervised instruction fine-tuning and linear RoPE scaling.
    \item \texttt{longchat-7b-16k} \footnote{https://huggingface.co/lmsys/longchat-7b-16k} is trained by fine-tuning Llama-7b on user-shared conversations collected from ShareGPT, using the condensing rotary embedding technique.
    \item \texttt{longchat-7b-32k} \footnote{https://huggingface.co/lmsys/longchat-7b-v1.5-32k} is the 32k version of vicuna-v1.5-16k.
    \item \texttt{chatglm3-6b-32k} \footnote{https://huggingface.co/THUDM/chatglm3-6b-32k} is ChatGLM 3 with 32k context window.
    \item \texttt{gpt-3.5-16k} and \texttt{gpt-4-128k} \footnote{https://platform.openai.com/docs/models/gpt-4-and-gpt-4-turbo} are gpt-3.5-turbo-1106 and gpt-4-1106-preview models provided by OpenAI.
\end{itemize}

Three retrievers in our experiments are built from Llama-index \cite{Liu_LlamaIndex_2022} library. Our experiments run on four A5000 GPUs with 25G memory space.

\subsection{Retriever Evaluation}
\label{ap:retri_eval}

For models using RAG method, retrievers use the query $Q_i$ to retrieve the top $u$ relevant text chunks with size $l$, i.e. $\mathrm{X}_i=\{x_{i,j}: j \in [1,u]\}$, as shown in Figure \ref{fig:pipeline_models}. These chunks $\mathrm{X}$ and QA pairs are then fed into LLMs to get the final response. Recall that the gold article and timestamp for $Q_i$ are $A_{i,gold}$ and $t_{i,gold}$. Each text chunk also has its own timestamp $t_{i,j}$ and is given to LLMs alongside $x_{i,j}$.

In TLB-detail, we use two metric, \textit{Acc\_Doc} and \textit{Acc\_Date}, which shows in how many questions the retriever finds the gold articles and timestamps respectively. In TLB-order, we use \textit{Acc\_Dates} which shows in how many questions the retriever locates all the three gold timestamps $\mathrm{T}_C =\{t_{C_r}: r \in [1,R]\}$. Their definitions are shown in Eq.\ref{eq:Acc_Doc}, Eq.\ref{eq:Acc_Date} and Eq.\ref{eq:Acc_Dates} respectively, where $N$ is the total number of questions, $\mathrm{I}(\cdot)$ is the sign function,  $\mathrm{T}_{i,X}$ and $\mathrm{T}_{i,C}$ are the sets of timestamps of retrieved text chunks and choices for the query $Q_i$ respectively. Note that $R=u=3$, indicating that the number of elements in $\mathrm{T}_{i,X}$ and $\mathrm{T}_{i,C}$ are the same.

\begin{small} 
\begin{equation}
\setlength\abovedisplayskip{2pt}
\setlength\belowdisplayskip{2pt}
Acc\_Doc = \frac{1}{N}\sum_{i=1}^{N}\mathrm{I}(\sum_{j=1}^{u}\mathrm{I}(x_{i,j} \in A_{i,gold})>0)
\label{eq:Acc_Doc}
\end{equation}
\end{small}

\begin{small} 
\begin{equation}
\setlength\abovedisplayskip{2pt}
\setlength\belowdisplayskip{2pt}
Acc\_Date = \frac{1}{N}\sum_{i=1}^{N}\mathrm{I}(\sum_{j=1}^{u}\mathrm{I}(t_{i,j} = t_{i,gold})>0)
\label{eq:Acc_Date}
\end{equation}
\end{small}

\begin{small} 
\begin{equation}
\setlength\abovedisplayskip{2pt}
\setlength\belowdisplayskip{2pt}
Acc\_Dates = \frac{1}{N}\sum_{i=1}^{N} \mathrm{I}(\mathrm{T}_{i,X} = \mathrm{T}_{i,C})
\label{eq:Acc_Dates}
\end{equation}
\end{small}

\subsection{Truncation of Long Input}
\label{ap:inpt_limit}

For LLM with long context window, if the input exceeds the limit of its context window, some articles are discard follwing the rule below, except those on the gold timestamp(s). Recall that news articles accessible to models are $\mathcal{A}_{n-1} =\{\mathrm{A_k}: k \in [1,n-1] \}$ without those on $t_n$.

\noindent{\textbf{TLB-detail.}} We normally discard the articles one by one from the last accessible timestamp $t_{n-1}$, until the input fits into the context window. However, there are chances that articles between $t_1$ and $t_{gold}$ exceed the input limit. In this case, we discard articles from the first timestamp $t_1$. When the articles between $t_1$ and $t_{gold}$ and between $t_{gold}$ and $t_{n-1}$ both exceed the input limit, we discard articles from $t_1$ and $t_{n-1}$ at the same time.

\noindent{\textbf{TLB-order.}} The ranking problem in TLB-order has three choices with three timestamps as part of the ground truth, i.e. $t_{1,gold} < t_{2,gold} < t_{3,gold}$. We normally discard the articles one by one from $t_{n-1}$ to $t_{3,gold}$ until fitting into the context window. When not working, we discard those from $t_{1}$ to $t_{1,gold}$. However, there are chances that articles between $t_{1,gold}$ and $t_{3,gold}$ exceed the input limit. In this case, we randomly sample articles between $t_{1,gold}$ and $t_{3,gold}$, but not in $t_{2,gold}$, one by one, until fitting into the context window.

\noindent{\textbf{TLB-forecast.}} We discard the articles one by one from the first timestamp $t_{1}$ to $t_{n-1}$, until the input fits into the context window.

\section{Prompt Strategy}
\label{ap:prompt_strategy_all}

\subsection{Outline Extraction}
\label{ap:outline_prompt}
The few-shot prompt for key point extraction is in Table \ref{table:prompt_point_extr}.

\begin{table*}[h!]  \small
\setlength\tabcolsep{2pt}
\centering
\renewcommand{\arraystretch}{1}
\begin{tabular}{p{16cm}}
 \specialrule{0.08em}{3pt}{3pt}
 You are an expert in extracting key contents from articles. \\
 \textbf{[Rules:]} Please extract the key points from the article with the following rules:\\
 1. Points should be independent from each other and have little overlaps.\\
 2. Points should be concise, accurate and complete, especially for numbers, names and dates.\\
 3. If points discuss events happened over one month ago, please discard them and keep those discussing events that just happened.\\
 4. Basically NO "he, she, they, it, them, etc" are allowed. Please clearly write out the entity you are referencing in the point.\\
 5. You are not allowed to start with any of the phrases: the article discusses, the article shows, the article emphasizes, the article discusses, the speaker says, the speaker discusses, the author mentions, etc.\\
 \textbf{[Example:]} Here are several examples of extracting key points from articles. Note that the articles in different examples are irrelevant.\\
 \textbf{Example 1:} \\
 Article: Islamic Jihad has threatened military action against Israel if Palestinian prisoner Hisham Abu Hawash, who is on a hunger strike, dies. Abu Hawash has been on a hunger strike for more than four months in protest of his detention without trial. Islamic Jihad spokesman Daoud Shihab said that "all options are on the table" and that the group is in urgent contact with Egyptian mediators to prevent an escalation. Senior Islamic Jihad official Khaled al-Batash said that if Abu Hawash dies, there would be a joint response from all factions in Gaza, including Hamas' military wing. Dozens of protests and strikes are taking place in Palestinian cities in solidarity with Abu Hawash, including a planned strike on Tuesday in his hometown of Dura.\\
 Key Points:  \\
 * Islamic Jihad has threatened military action against Israel if Palestinian prisoner Hisham Abu Hawash dies. \\
* Islamic Jihad is in urgent contact with Egyptian mediators to prevent an escalation. \\
* Islamic Jihad would start a joint response from all factions in Gaza, including Hamas' military wing if Palestinian prisoner Hisham Abu Hawash dies. \\
* Protests and strikes take place in Palestinian cities in solidarity with Palestinian prisoner Hisham Abu Hawash. \\
 \textbf{Example 2:} \\
 Article:  \\
 Islamic Jihad has threatened military action against Israel if Palestinian prisoner Hisham Abu Hawash, who is on a hunger strike, dies. Abu Hawash has been on a hunger strike for more than four months in protest of his detention without trial. Islamic Jihad spokesman Daoud Shihab said that "all options are on the table" and that the group is in urgent contact with Egyptian mediators to prevent an escalation. Senior Islamic Jihad official Khaled al-Batash said that if Abu Hawash dies, there would be a joint response from all factions in Gaza, including Hamas' military wing. Dozens of protests and strikes are taking place in Palestinian cities in solidarity with Abu Hawash, including a planned strike on Tuesday in his hometown of Dura.  \\
 Key Points:   \\
 * Islamic Jihad has threatened military action against Israel if Palestinian prisoner Hisham Abu Hawash dies. \\
* Islamic Jihad is in urgent contact with Egyptian mediators to prevent an escalation. \\
* Islamic Jihad would start a joint response from all factions in Gaza, including Hamas' military wing if Palestinian prisoner Hisham Abu Hawash dies. \\
* Protests and strikes take place in Palestinian cities in solidarity with Palestinian prisoner Hisham Abu Hawash. \\
 \textbf{Example 3:} \\
 Article:  \\
 Israel has announced that it is gradually reopening its embassy in Jordan after a shutdown prompted by a deadly shooting in the embassy's vicinity last year. The shooting, which was carried out by a security guard for the Israeli embassy, resulted in the death of two Jordanian workers, including one who had stabbed the guard with a screwdriver. The incident sparked widespread anger in Jordan, and the Jordanian government refused to allow the embassy staff to return until Israel opened a serious investigation and offered an apology. In January, Israel reportedly apologized and agreed to compensate the families of the victims, and the conditions for reopening the embassy were met. The embassy staff received a hero's welcome from Israeli Prime Minister Benjamin Netanyahu, who was accompanied by the Israeli ambassador.  \\
 Key Points:   \\
 * Israel has announced to gradually reopen Isreal's embassy in Jordan after a shutdown. \\
* One Jordanian worker stabbed a security guard for the Israeli embassy with a screwdriver, and the guard shot two Jordanian workers to death. \\
* The Jordanian government refused to allow the security guard to return until Israel opened a serious investigation and offered an apology. \\
* Israel reportedly apologized and agreed to compensate the families of the victims to meet the conditions for reopening the Israeli embassy in Jordan. \\
* The security guard received a hero's welcome from Israeli Prime Minister Benjamin Netanyahu. \\
 \textbf{[New Article:]} Given the above rules and examples, please extract the key points of the following article and output them in the same way as examples. \\
 Article: \{\textbf{\textcolor{OliveGreen}{Summary}}\} \\
 \textbf{[Output:]} Key Points: \\
 \specialrule{0.08em}{2pt}{2pt}
\end{tabular}
\caption{Few-shot prompt for key point extraction. The daily summary to be split enters \textbf{\textcolor{OliveGreen}{Summary}}. We call daily summary as article in the prompt in case of misleading LLMs.}
\label{table:prompt_point_extr}
\end{table*}

\subsection{Dataset Construction}
\label{ap:prompt_generation}
% Generate-then-verify Paradigm
The few-shot prompts for QA generation in TLB-detail and TLB-forecast are in Table \ref{table:prompt_detailQA_gen} and Table \ref{table:prompt_foreQA_gen} respectively. The few-shot prompt for misleading choices generation is in Table \ref{table:prompt_choices_gen}.

The prompt templates for verifying \textit{Evidence}, \textit{Forecasting}, and \textit{Storytelling} and \textit{Temporal} are in Table \ref{table:prompt_veri_evidence}, Table \ref{table:prompt_veri_forec} and Table \ref{table:prompt_veri_story_temp} respectively.

\begin{table*}[h!]  \small
\setlength\tabcolsep{2pt}
\centering
\renewcommand{\arraystretch}{1}
\begin{tabular}{p{16cm}}
 \specialrule{0.08em}{3pt}{3pt}
\textbf{[Rules:]} Article: \{\textbf{\textcolor{OliveGreen}{Article}}\} \\
Given the above article, please generate one question along with its answer. You should follow the instructions below: \\
1. The question should be around the key point "\{\textbf{\textcolor{MidnightBlue}{Point}}\}" and come from the above article as well. \\
2. The question should be unambiguous and challenging, avoiding simple string matching. NO sub-questions allowed. \\
3. The question should be answerable based only on the text of the above article. \\
4. You should avoid the following question types: questions that require numerical reasoning (this is not a math test); questions that require substantial world knowledge; questions that require the reader to speculate. \\
5. The answer MUST be short and concise, avoiding using redundant words or repeating the information in the question. \\
6. You should output the question and its answer without any other explanation, such as "Question: xxx? Answer: xxx." \\
\textbf{[Example:]} Here are some examples showing the writing style. NOTE that the content of the examples are irrelevant to the question you will generate. \\
* Question: What does Holger von Neuhoff say about the bottled message? Answer: It is the oldest message found along with the bottle he has ever encountered \\
* Question: Who first stated that the polygraph might not be reliable?? Answer: The psychologist William Martson \\
* Question: Where did Richard Platz want the postcard to end up? Answer: At a museum
* Question: When are police stations expected to start using the new lie detection method? Answer: Once it reaches an accuracy of at least 70\% \\
* Question: What is a challenge working children face in regards to attending school, according to al-Mamun? Answer: It can be hard for them to assimilate to the school environment \\
 \textbf{[Output:]} Now please write a question following the instructions and examples above. You should output the question along with its answer, in the format of "Question: xxx? Answer: xxx.". NOTE that the answer should be as short as possible. \\
 \specialrule{0.08em}{2pt}{2pt}
\end{tabular}
\caption{Few-shot prompt for QA generation of MCQ in TLB-detail. \textbf{\textcolor{MidnightBlue}{Point}} and \textbf{\textcolor{OliveGreen}{Article}} are a key point and article with the same timestamp. The examples are from \citet{berzak-etal-2020-starc}.}
\label{table:prompt_detailQA_gen}
\end{table*}

\begin{table*}[h!]  \small
\setlength\tabcolsep{2pt}
\centering
\renewcommand{\arraystretch}{1}
\begin{tabular}{p{16cm}}
 \specialrule{0.08em}{3pt}{3pt}
\textbf{[Time Setup:]} Imagine the scenario: Today is \{\textbf{\textcolor{Bittersweet}{Day}}\}. The article provided has just been published. \\
\textbf{[Rules:]} Article: \{\textbf{\textcolor{OliveGreen}{Article}}\}. Publishing date: \{\textbf{\textcolor{Bittersweet}{Day}}\} \\
Please generate one forecasting question about the above article, along with its answer. You should follow the instructions below: \\
1. The question should be around the key point "\{\textbf{\textcolor{MidnightBlue}{Point}}\}" and come from the above article. \\
2. The question must be guessable, but not answerable until  \{\textbf{\textcolor{Bittersweet}{Day}}\}.  \\
3. The question should start with one of the following phrases: "What will", "Who will", "Where will", "Which country will", "Why will", "How much", "How will", "How many". \\
4. There must be a time element in the question. It can be phrases like "In  \{\textbf{\textcolor{Bittersweet}{Day}}\} ...", "After  \{\textbf{\textcolor{Bittersweet}{Day}}\}, ...", "... in  \{\textbf{\textcolor{Bittersweet}{Day}}\}?". However, you are NOT allowed to use "before" in the question, as remember the question should be able to be answered without information from the day the article was published. \\
5. You should avoid: questions that require numerical reasoning; questions that require substantial world knowledge. \\
6. The answer MUST be short and concise, avoiding using redundant words or repeating the information in the question. \\
7. The question must be grammatically correct and contain the information required to answer. NO "he, she, they, it, them, etc" allowed. Please clearly write out the entity you are referencing in the foercasting question. \\
\textbf{[Example:]} Here are some examples showing the writing style. NOTE that the content of the examples are irrelevant to the question you will generate. \\
* Question: What will Belinda Carlisle want to be by 2019-09-01? Answer: Travel Agent \\
* Question: Who will visit Pittsburgh for first 2020 campaign rally in 2019-04-12? Answer: Joe Biden \\
* Question: Where will the Glasgow derby be played in 2021-05-01? Answer: Scotland \\
* Question: What will be M\&S's response after their shares fall in 2016-03-24? Answer: They will focus on the goal and aim to regenerate the business within the next 5 years \\
* Question: What will Trump say that will happen to the economy if he's not reelected in 2017-08-13? Answer: The economy will tank \\
 \textbf{[Output:]} Now please write a question following the instructions and examples above. You should output the question along with its answer, in the format of "Question: xxx? Answer: xxx.". \\
 \specialrule{0.08em}{2pt}{2pt}
\end{tabular}
\caption{Few-shot prompt for QA generation of MCQ in TLB-forecast. \textbf{\textcolor{MidnightBlue}{Point}} and \textbf{\textcolor{OliveGreen}{Article}} are a key point and article on \textbf{\textcolor{Bittersweet}{Day}}. \textbf{\textcolor{Bittersweet}{Day}} is the last timestamp of TCE. The instruction is borrowed from \citet{jin-etal-2021-forecastqa}, and examples also from \citet{jin-etal-2021-forecastqa}.}
\label{table:prompt_foreQA_gen}
\end{table*}

\begin{table*}[h!]  \small
\setlength\tabcolsep{2pt}
\centering
\renewcommand{\arraystretch}{1}
\begin{tabular}{p{16cm}}
 \specialrule{0.08em}{3pt}{3pt}
\textbf{[Rules:]}  Background 1: \{\textbf{\textcolor{OliveGreen}{Article 1}}\}. Background 2: \{\textbf{\textcolor{JungleGreen}{Article 2}}\} \\
Given above two backgrounds, please generate three noising answers to the question "\{\textbf{\textcolor{Violet}{Question}}\}", whose correct answer is "\{\textbf{\textcolor{VioletRed}{Answer}}\}". Name the three noising answers as (b), (c) and (d) respectively. You should follow the instructions below: \\
1. (b), (c) and (d) must share the similar wording and length with the correct answer "\{\textbf{\textcolor{VioletRed}{Answer}}\}". \\
2. The four answers must be essentially different and contradictory.  \\
3. Answer (b) is incorrect and reflects a misunderstanding of Background 1. (b) should not repeat the correct answer "\{\textbf{\textcolor{VioletRed}{Answer}}\}". \\
4. Answer (c) is incorrect and comes from Background 2. \\
5. Answer (d) is incorrect and has no support in neither of the backgrounds. (d) may refer to general world knowledge. \\
6. While (c) and (d) should all be unambiguously incorrect, they should also make sense and be plausible answers to the question. \\
7. (c) and in some cases (b) could be correct (in part or fully) as a fact but not correct as an answer to the question. It's also fine for (c) to be an incorrect fact as long as it has textual support in Background 2. \\
\textbf{[Example:]} Here are examples showing the output format. This example is NOT related to the noising answers you will generate. \\
\textbf{Question:} \\
Who threw the bottle into the Baltic Sea? \\
Correct Answer: \\
Angela Erdmann. \\
Nosing Answers: \\
(b) Angela Erdmann's grandfather. \\
(c) A museum worker. \\
(d) A fisherman. \\
\textbf{Question:} \\
What does Erdmann want to add to the bottle exhibit? \\
Correct Answer: \\
Pictures of the bottled message's author \\
Nosing Answers: \\
(b) A deciphered copy of the text \\
(c) A photo that depicts a young man throwing a bottle into the sea \\
(d) Excerpts from a book written by her grandfather \\
\textbf{Question:} \\
Where does Dunamn believe the athletic abilities of adults are derived from? \\
Correct Answer: \\
The month in which they were born in \\
Nosing Answers: \\
(b) The opportunities offered by UK Sport during their youth \\
(c) Primarily from their innate genetics \\
(d) A combination of multiple different factors \\
\textbf{Question:}  \\
What is a challenge working children face in regards to attending school, according to al-Mamun? \\
Correct Answer: \\
It can be hard for them to assimilate to the school environment \\
Nosing Answers: \\
(b) After they stop working, they miss their friends from the factory \\
(c) SOHAY's classes are intended for parents and employers, not children \\
(d) They don't have enough preparation for the level of learning \\
\textbf{Question:}  \\
When are police stations expected to start using the new lie detection method? \\
Correct Answer: \\
Once it reaches an accuracy of at least 70\% \\
Nosing Answers: \\
(b) Within 10 years \\
(c) Once it is able to track the movements of the entire body \\
(d) It is already in use in many police stations \\
 \textbf{[Output:]} Now please generate three noising answers to the question, given the above backgrounds, instructions and examples. DO NOT output the backgrounds, the question or any other explanations. \\
Question:  \\
\{\textbf{\textcolor{Violet}{Question}}\}.  \\
Correct Answer:  \\
\{\textbf{\textcolor{VioletRed}{Answer}}\}.  \\
Nosing Answers: \\
 \specialrule{0.08em}{2pt}{2pt}
\end{tabular}
\caption{Few-shot prompt for misleading choices generation of MCQ in TLB-detail and TLB-forecast. \textbf{\textcolor{OliveGreen}{Article 1}} is the article used for generating \textbf{\textcolor{Violet}{Question}} and \textbf{\textcolor{VioletRed}{Answer}}. \textbf{\textcolor{JungleGreen}{Article 2}} is a random article on another random timestamp.
The instruction and examples are from \citet{berzak-etal-2020-starc}.}
\label{table:prompt_choices_gen}
\end{table*}

\begin{table*}[h!]  \small
\setlength\tabcolsep{2pt}
\centering
\renewcommand{\arraystretch}{1}
\begin{tabular}{p{16cm}}
 \specialrule{0.08em}{3pt}{3pt}
\textbf{[Rules:]} 
Article:  \\
\{\textbf{\textcolor{OliveGreen}{Article}}\}.  \\
Question:  \\
\{\textbf{\textcolor{Violet}{Question}}\}.  \\
Answer:  \\
\{\textbf{\textcolor{VioletRed}{Answer}}\}. \\
Given the above articles, please check if the answer is correct to the question with 100\% certainty. You should follow the instructions below: \\
1. You should first find the relevant sentences from the above article. \\
2. You should then reason out the answer to the above question step by step. \\
3. Finally, you should compare your answer with the above one.  \\
 \textbf{[Output:]} If the above answer is the same as the one you got, please output "The given answer is correct." along with one original sentence that supports the answer the most strongly; otherwise, output "The given answer may be wrong." along with one original sentence that rejects the answer the most strongly. \\
 \specialrule{0.08em}{2pt}{2pt}
\end{tabular}
\caption{Prompt template for verifying \textit{Evidence}.}
\label{table:prompt_veri_evidence}
\end{table*}

\begin{table*}[h!]  \small
\setlength\tabcolsep{2pt}
\centering
\renewcommand{\arraystretch}{1}
\begin{tabular}{p{16cm}}
 \specialrule{0.08em}{3pt}{3pt}
\textbf{[Rules:]}Please verify the question. \\
Question Asked: \{\textbf{\textcolor{Violet}{Question}}\} \\
Note: The above question and its answer come from one article on \{\textbf{\textcolor{Bittersweet}{Day}}\}.
Situation: In order to answer the above question you are given access to all news articles published before \{\textbf{\textcolor{Bittersweet}{Day}}\}. \\
\textbf{Task Context:} You can imagine going back in time to one day before \{\textbf{\textcolor{Bittersweet}{Day}}\}, and on this day you are being posed the question above, while having access to the articles stated in the situation provided. \\
\textbf{Q1:} Do you think a person (could be anyone, even an expert in the field) would you be able to make an educated guess as to what the answer to this question is, given the provided situation? \\
A. Yes, the person would be able to make an educated guess as to what the answer to this question is. \\
B. No, the person would not be able to make an educated guess as to what the answer to this question is. \\
C. I'm not sure/I can't answer/Other \\
\textbf{Q2:} Do you think a person (could be anyone, even an expert in the field) would be able to find an article (or many) published before \{\textbf{\textcolor{Bittersweet}{Day}}\} that answers the question with 100\% certainty? \\
Note: We don't mean a guess, but rather the article would have a passage that either by itself or with the help of other passages from other articles (all published before \{\textbf{\textcolor{Bittersweet}{Day}}\}) would directly answer this question. \\
A. Yes, the person would find article(s) from before \{\textbf{\textcolor{Bittersweet}{Day}}\} that would directly answer this question. \\
B. No, the person would need information from article(s) from \{\textbf{\textcolor{Bittersweet}{Day}}\} or after to directly answer this question. \\
C. I'm not sure/I can't answer/Other \\
\textbf{[Output:]}Please output your answer to Q1 and Q2, in the format of "Q1: x. Q2: x". \\
 \specialrule{0.08em}{2pt}{2pt}
\end{tabular}
\caption{Prompt template for verifying \textit{Forecasting}.}
\label{table:prompt_veri_forec}
\end{table*}

\begin{table*}[h!]  \small
\setlength\tabcolsep{2pt}
\centering
\renewcommand{\arraystretch}{1}
\begin{tabular}{p{16cm}}
 \specialrule{0.08em}{3pt}{3pt}
\textbf{[Rules:]}Below are key points presenting a storyline. Please verify this storyline. \\
\{\textbf{\textcolor{MidnightBlue}{Points for Ranking}}\} \\
Q1: Do you think the above key points are arranged in a chronological order? \\
A. Yes, the above key points are apparently arranged in a chronological order. \\
B. No, swapping some of them can make the storyline more chronological. \\
C. I'm not sure/I can't answer/Other \\
Q2: Do you think each of the above key points represents a event that just happened or is happening? \\
A. Yes, they all represent the events that just happened or is happening. \\
B. No, some of them discuss the static content of certain documents, someone's view or events that may happen in the future and/or happened before. \\
C. I'm not sure/I can't answer/Other \\
\textbf{[Output:]}Please output your answer to Q1 and Q2, in the format of "Q1: x. Q2: x". \\
 \specialrule{0.08em}{2pt}{2pt}
\end{tabular}
\caption{Prompt template for verifying \textit{Storytelling} and \textit{Temporal}.}
\label{table:prompt_veri_story_temp}
\end{table*}

\subsection{Evaluation}
\label{ap:prompt_eval}
The prompt templates for evaluation in TLB-detail, TLB-order and TLB-forecast are in Table \ref{table:prompt_eval_detail}, Table \ref{table:prompt_eval_order}, and Table \ref{table:prompt_eval_fore} respectively.

\begin{table*}[h!]  \small
\setlength\tabcolsep{2pt}
\centering
\renewcommand{\arraystretch}{1}
\begin{tabular}{p{16cm}}
 \specialrule{0.08em}{3pt}{3pt}
\textbf{[System Message:]} You're an expert in answering multiple choice questions. And you will never refuse to answer any question. \\
\textbf{[Rule:]} \{\textbf{\textcolor{OliveGreen}{Context}}\} \\
Given the above articles, please select one of the option that is the most appropriate for the question below. Note that you will never refuse to answer a question. \\
You should output your answer like 'X. x.' WITHOUT anything else, where 'x' is the choice's letter. \\
Question: \\
\{\textbf{\textcolor{Violet}{Question}}\} \\
Choices: \\
\{\textbf{\textcolor{VioletRed}{Candidate Choices}}\} \\
\textbf{[Output:]} Your answer: \\
 \specialrule{0.08em}{2pt}{2pt}
\end{tabular}
\caption{Prompt template for evaluation in TLB-detail. \textbf{\textcolor{OliveGreen}{Context}} consists of retrieved text chunks/articles and their corresponding timestamps.}
\label{table:prompt_eval_detail}
\end{table*}

\begin{table*}[h!]  \small
\setlength\tabcolsep{2pt}
\centering
\renewcommand{\arraystretch}{1}
\begin{tabular}{p{16cm}}
 \specialrule{0.08em}{3pt}{3pt}
\textbf{[System Message:]} You are an expert in ordering several sentences to form a chronological storyline. And you will never refuse to order any choice. \\
\textbf{[Rule:]}  \{\textbf{\textcolor{OliveGreen}{Context}}\} \\
Given the above articles, please order the following choices to form a chronological storyline. Note that you will never refuse to order any choice. \\
You should output your answer like  'x,x,x.' WITHOUT anything else, where 'x' is the choice's letter. \\
Choices: \\
\{\textbf{\textcolor{VioletRed}{Candidate Choices}}\} \\
\textbf{[Output:]} Your answer: \\
 \specialrule{0.08em}{2pt}{2pt}
\end{tabular}
\caption{Prompt template for evaluation in TLB-order. \textbf{\textcolor{OliveGreen}{Context}} consists of retrieved text chunks/articles and their corresponding timestamps.}
\label{table:prompt_eval_order}
\end{table*}

\begin{table*}[h!]  \small
\setlength\tabcolsep{2pt}
\centering
\renewcommand{\arraystretch}{1}
\begin{tabular}{p{16cm}}
 \specialrule{0.08em}{3pt}{3pt}
\textbf{[System Message:]} You're an expert in forecasting events. You can find out what will happen next given the latest information, even if you are not with 100\% certainty. And you will never refuse to answer a forecasting question. \\
\textbf{[Rule:]}  \{\textbf{\textcolor{OliveGreen}{Context}}\} \\
Given the above articles, please select the option that is the most likely to be the correct answer the the question. Note that you will never refuse to answer a forecasting question, even if without 100\% certainty. \\
You should output your answer like 'X. x.' WITHOUT anything else, where 'x' is the choice's letter. \\
Question: \\
\{\textbf{\textcolor{Violet}{Question}}\} \\
Choices: \\
\{\textbf{\textcolor{VioletRed}{Candidate Choices}}\} \\
\textbf{[Output:]} Your answer: \\
 \specialrule{0.08em}{2pt}{2pt}
\end{tabular}
\caption{Prompt template for evaluation in TLB-forecast. \textbf{\textcolor{OliveGreen}{Context}} consists of retrieved text chunks/articles and their corresponding timestamps.}
\label{table:prompt_eval_fore}
\end{table*}

\end{document}